\documentclass[final]{cvpr}

\usepackage{times}
\usepackage{epsfig}
\usepackage{graphicx}
\usepackage{amsmath}
\usepackage{amssymb}
\usepackage{adjustbox}
\usepackage{multirow}
\usepackage{rotating}
\usepackage{tabularx} 
\usepackage{booktabs}
\usepackage{amsthm}

\newcommand{\vI}{\mathbf{I}}
\newcommand{\vA}{\mathbf{A}}
\newcommand{\vC}{\mathbf{C}}
\newcommand{\vB}{\mathbf{B}}

\newcommand{\vt}{\mathbf{t}}
\newcommand{\vp}{\mathbf{p}}
\newcommand{\vX}{\mathbf{X}}
\newcommand{\vY}{\mathbf{Y}}
\newcommand{\vR}{\mathbf{R}}

\newcommand{\bX}{\mathrm{X}}
\newcommand{\bY}{\mathrm{Y}}
\newcommand{\bZ}{\mathrm{Z}}
\newcommand{\bW}{\mathrm{W}}
\newcommand{\R}{{\rm I} \! {\rm R}}

\newcommand{\vF}{\vec{\mathcal{F}}}
\newcommand{\vG}{\vec{\mathcal{G}}}
\newcommand{\vD}{\mathbf{D}}
\newcommand{\vd}{\mathbf{d}}
\newcommand{\vL}{\mathbf{I}}

\newcommand{\calL}{\mathcal{L}}
\newcommand{\calS}{\mathcal{S}}
\newcommand{\calI}{\mathcal{I}}

\newcommand{\calT}{\mathcal{T}}

\usepackage[pagebackref=true,breaklinks=true,colorlinks,bookmarks=false]
{hyperref}

\def\cvprPaperID{1427} 
\def\confYear{CVPR 2021}
\begin{document}

\title{Dual Pixel Exploration: Simultaneous Depth Estimation and Image
Restoration}

\author{Liyuan Pan\thanks{Equal contribution} ,  Shah Chowdhury\footnotemark[1] , Richard Hartley, Miaomiao Liu, Hongguang Zhang,~and Hongdong Li\\ 
Australian National University, Canberra, Australia \ \ \ Australian Centre for Robotic Vision \\
\tt\small{\{liyuan.pan,Shah.Chowdhury\}}@anu.edu.au
}

\maketitle

\begin{abstract}
The dual-pixel (DP) hardware works by splitting each pixel in half and creating
an image pair in a single snapshot. Several works estimate depth/inverse depth by
treating the DP pair as a stereo pair. However, dual-pixel disparity only occurs
in image regions with the defocus blur. The heavy defocus blur in DP pairs
affects the performance of matching-based depth estimation approaches. 
Instead of removing the blur effect blindly, we study the formation of the DP
pair which links the blur and the depth information. In this paper, we propose a
mathematical DP model which can benefit depth estimation by the blur. 
These explorations motivate us to propose an end-to-end \textbf{DDDNet}
(DP-based Depth and Deblur Network) to jointly estimate the depth and
restore the image. Moreover, we define a reblur loss, which reflects the
relationship of the DP image formation process with depth information, to
regularise our depth estimate in training. To meet the requirement of a large
amount of data for learning, we propose the first DP image simulator which
allows us to create datasets with DP pairs from any existing RGBD dataset.
As a side contribution, we collect a real dataset for further research.
Extensive experimental evaluation on both synthetic and real datasets shows that
our approach achieves competitive performance compared to state-of-the-art
approaches.
\end{abstract}

\section{Introduction}
\begin{figure}[t]
    \centering
    \begin{tabular}[t]{ccccc}
    \hspace{-0.451 cm}
    \includegraphics[width=0.160\textwidth]{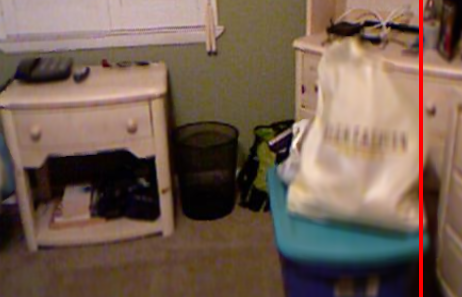}
    \hspace{-0.451 cm}
    &\includegraphics[width=0.160\textwidth]{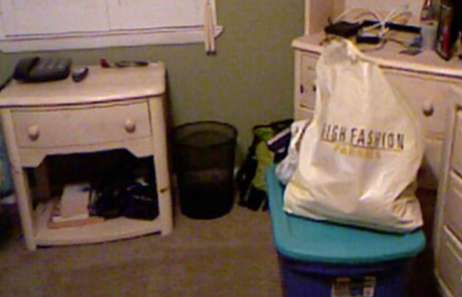}
    \hspace{-0.451 cm}
    &\includegraphics[width=0.160\textwidth]{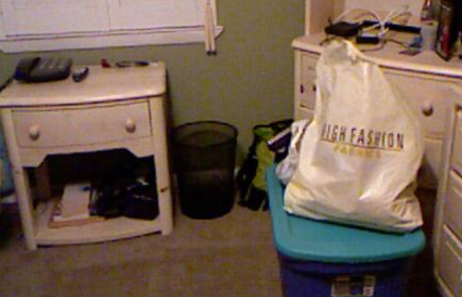}\\
    \hspace{-0.451 cm}
    \includegraphics[width=0.160\textwidth]{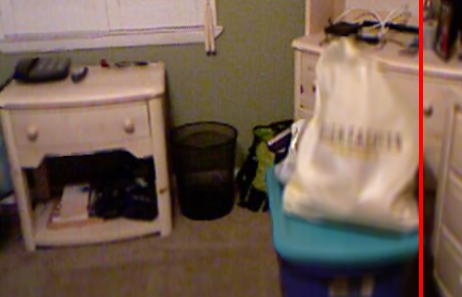}
    \hspace{-0.451 cm}
    &\includegraphics[width=0.160\textwidth]{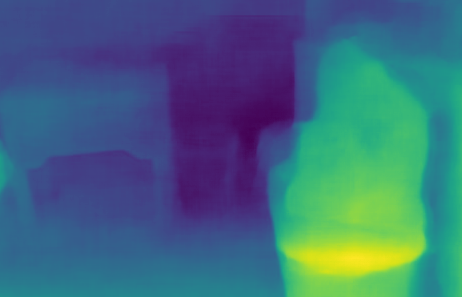}
    \hspace{-0.451 cm}
    &\includegraphics[width=0.160\textwidth]{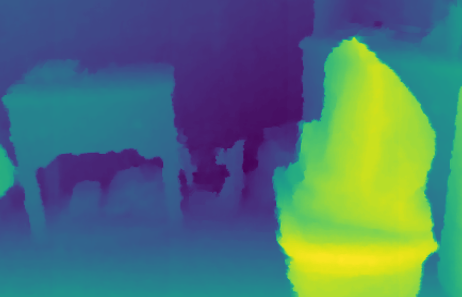}\\
    \hspace{-0.451 cm}
    (a) Inputs
    \hspace{-0.451 cm} 
    & (b) Ours
    \hspace{-0.451 cm}
    & (c) GT\\
    \end{tabular}
    \caption{\it The pipeline of our approach. (a) The input of our {\bf
DDDNet}. A red line in each DP image is drawn to indicate a small shift between
the left and right sub-aperture views. (b) Our deblurred image and estimated
inverse depth map. (c) The Ground-Truth (GT) sharp image and inverse depth map. The
input DP pair is from a real-world DP sensor, or generated by our simulator.
    }
    \label{fig:DPmodel-pipe}
    \vspace{-2mm}
\end{figure}
The Dual-Pixel (DP) sensor
has been used by DSLR (digital single-lens reflex camera)
and smartphone cameras to aid focusing. 
Though the DP sensor is designed for auto-focus~
\cite{sliwinski2013simple,jang2015sensor,herrmann2020learning}, it is used in
applications such as, depth estimation~
\cite{garg2019learning,punnappurath2020modeling,zhang20202}, defocus
deblurring~\cite{abuolaim2020defocus}, reflection
removal~\cite{punnappurath2019reflection}, and shallow Depth-of-Field (DoF)
images synthesis~\cite{wadhwa2018synthetic}. In this paper, we model the
imaging
process of the DP sensor theoretically, and show its effectiveness for
simultaneous depth
estimation and image deblurring.

A DP camera simultaneously captures two images, one formed from light
rays passing through the right half of the aperture, and one from those passing
through the left half
of the aperture.  Because of the displacement of these two half-apertures,
the two images form a stereo pair. The primary role of DP cameras is to 
enable auto-focus.  The set of points in the world that are exactly in focus
forms a plane parallel to the lens (perpendicular to the axis of the camera),
and points that lie on this plane are both in focus, and imaged with
zero-disparity in the two
images.  Points that lie further away from the camera are imaged with positive
disparity
in the two images and those that are closer have negative disparity.
By detecting and modelling these disparities,
it is possible to compute scene depth, as in standard stereo imaging.
(In auto-focus these disparities are used to move the sensor, or the lens, to
focus on any part of the scene.)

At the same time, points other than those lying on the in-focus plane will
be significantly out of focus, which raises the possibility of using 
depth-from-defocus
techniques to estimate depth in the scene.  This
out-of-focus effect is more significant than with ordinary
stereo pairs, making stereo matching potentially more challenging
for DP cameras.  However, because of the small base-line,
between the two half-apertures, occlusion effects are minimal.

Thus, computing depth from a DP camera can be seen as an
out-of-focus stereo estimation problem.  This paper aims to 
combine depth-from-disparity and depth-from-defocus approaches
in a single network, demonstrating the advantage of modelling
them both simultaneously.

To accomplish this,  we study the imaging process of a DP
pair and provide a mathematical model handling the intrinsic problem of {\em
depth from defocus blur} for a DP pair. Our method can jointly recover an
all-in-focus image and estimate the depth of a scene (see
Fig.~\ref{fig:DPmodel-pipe}). Our contributions are summarised as
follows:

\vspace{-2mm}
\begin{enumerate}
\itemsep-0.3em
\item We propose a theoretical DP model to explicitly define the relationship
between depth, defocus blur and all-in-focus image;
\item We design an end-to-end DP-based Depth and Deblur Network
(\textbf{DDDNet}) to jointly estimate the depth map and restore the sharp
image;
\item We formulate a reblur loss based on our DP model which is used to
regularise depth estimate in training;
\item We create a DP image simulator, which enables us to create DP datasets
from
any RGB-D dataset;
\item We collect a real dataset to stimulate further research.
\end{enumerate}
\vspace{-2mm}
Extensive experimental evaluation on both synthetic and real data shows that
our approach achieves competitive performance compared to state-of-the-art
approaches.

\section{Related Work}

For space reasons, we briefly review depth estimation from monocular, stereo,
DP, and defocus. 

\vspace{0.5mm}
{\noindent{\bf Monocular.}} Supervised monocular depth estimation methods
typically rely on large training datasets~
\cite{liu2015learning, fu2018deep}.
Self-supervised methods~\cite{saxena2007learning,
Tosi_2019_CVPR,lee2019big,monodepth2} often use estimated depths and camera
poses to generate synthesized images, serving as supervision signals.

\vspace{0.5mm}
{\noindent{\bf Stereo.}} To summarize, a stereo depth estimation
network~
\cite{hamzah2016literature,  vzbontar2016stereo, Nie_2019_CVPR, Duggal_2019_ICCV, Yin_2019_CVPR,aleotti2020reversing,xu2020aanet} 
often
contains: matching cost computation, cost aggregation, and disparity
optimization. Recently, several speed-up
methods~\cite{wang2019anytime,xu2020aanet} are available.

\vspace{0.5mm}
{\noindent{\bf Dual-pixel.}} A DP pair has defocus blur and tiny baseline. 
To tackle the tiny baseline, 
Wadhwa~\etal~\cite{wadhwa2018synthetic} utilize a small search size in block
matching. 
Garg~\etal~\cite{garg2019learning} present the first learning models for affine
invariant depth estimation to tackle affine ambiguity. 
Zhang~\etal~\cite{zhang20202} present a Du$^2$Net that uses a wide baseline
stereo camera to compensate for the DP camera. However, it is specifically
tailored for the `Google Pixel' (narrow aperture, small DoF). 
The above three methods ignored the blur cue that could in principle fill in the
parts where the disparity is imprecise (due to the tiny baseline). 
To tackle the defocus blur in a DP pair,
Abuolaim~\etal~\cite{abuolaim2020defocus} first propose a DPDNet that remove the
blur effect blindly. Recently, Punnappurath~\etal~
\cite{punnappurath2020modeling} use a point spread function (PSF) to model the
defocus blur and formulate the DP disparity by the PSF. However, the symmetry
assumption of the PSF (only holds for constant depth region) limits its
application for real-world scenario. 
Moreover, the method is time-consuming and has three steps (not end-to-end).
In contrast, we study the imaging process of a DP pair and give a mathematical
model jointly handling the intrinsic problem of {\em depth from defocus blur}
for a DP pair.

\vspace{0.5mm}
{\noindent{\bf Depth from defocus and deblurring.}} Depth from defocus has the
same geometric constraints as disparity but different physiological
constraints~\cite{ens1993investigation}. The estimated depth, also dubbed as
defocus map, is commonly used to guide the deblurring~\cite{chen2015blur,
suwajanakorn2015depth, xu2017estimating,
maximov2020focus,zhuo2011defocus,levin2007image,karaali2017edge}. 
DMENet~\cite{lee2019deep} proposes the first end-to-end CNN architecture to
estimate a spatially varying defocus map and use it for deblurring. 

\vspace{0.5mm}
{\noindent{\bf DP data.}} Only Canon and Google provide DP data to customers
though most DSLR and smartphone cameras have the DP sensor. Several
researchers~\cite{wadhwa2018synthetic, garg2019learning, zhang20202} use `Google
Pixel' to collect data. However, smartphone cameras use a fixed and narrow
aperture that cannot be adjusted. 
Canon is used  in~
\cite{abuolaim2020defocus,punnappurath2019reflection,punnappurath2020modeling} with different aperture sizes, but it is expensive.
Both of the two sensors are hard to get the associated GT information, such as
depth. In parallel, the rise of deep learning has led to an explosion of demands
for data. In this paper, we present the first DP simulator that synthesizes DP
images with GT, from any RGB-D data.

In summary, both the stereo cue and defocus cue provided by a DP pair benefit
the depth estimation. In this paper, we proposed a DP model in
Section~\ref{sec:DPmodel} and a {DDDNet} in Section~\ref{sec:DDDnet} to jointly
estimate the depth and restore the image.

\section{Dual-Pixel Image Formation}\label{sec:DPmodel}

In this section, we first discuss the formation of the two DP images and the
defocus blur in Section \ref{sec:model}. Then, in Section \ref{sec:rgbd}, we
present our DP model and explore how to synthesize DP images from the RGB-D
image. Based on the DP model, we build a DP simulator in Section \ref{sec:simu}.

\subsection{Model of a dual-pixel camera}\label{sec:model}

A DP camera can be modelled as an ordinary camera with a lens satisfying the
thin-lens model, in which two (or more) images $\vI_L$ and $\vI_R$ are
simultaneously captured. Here, the subscripts $_L$ and $_R$ denote the left and
right view separately.
In the model, the focal plane of the camera can be considered as consisting of
two identically placed focal planes, one of which captures light rays coming
from the left side of the lens, and the other captures light from the right side
of the lens. 
The two images, $\vI_L$ and $\vI_R$, can be considered as images taken by two
different (but coplanar) lens, corresponding to regions in the aperture of a
thin-lens. 

\begin{figure}[t]
    \centering
    \includegraphics[width=0.45\textwidth]{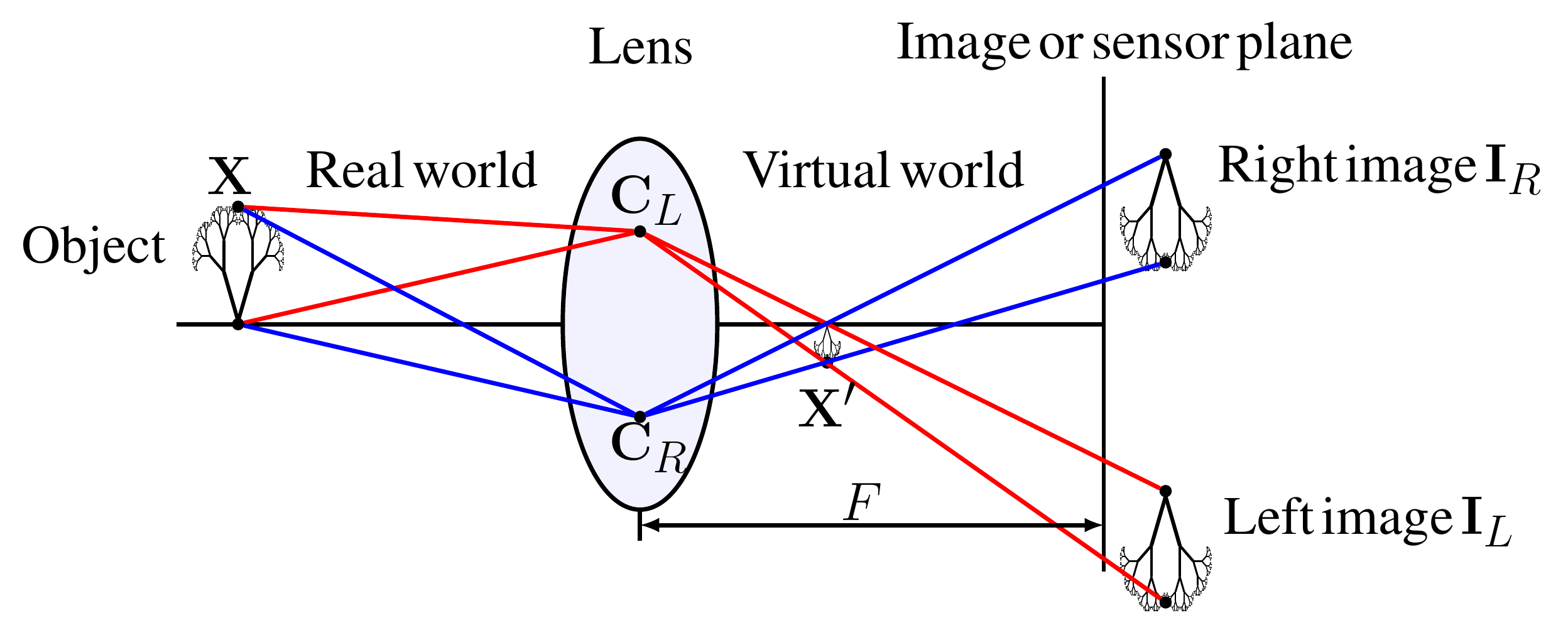}
    \vspace{-2mm}
    \caption{\it Formation of the two image $\vI_L$ and $\vI_R$ from two points
$\vC_L$ and $\vC_R$ on the lens plane may be thought of as pin-hole projections
(with centres $\vC_L$ and $\vC_R$ and focal-length $F$) of the virtual image
formed by the lens.}
    \vspace{-2mm}
    \label{fig:fig3}
\end{figure}

To be general, let it be assumed that the light for image $\vI_L$ passes through
a region $\vA_L$ of the lens, and that for image $\vI_R$ comes from a region
$\vA_R$.
Let us consider the case where $\vA_L$ and $\vA_R$ are infinitesimally small,
consisting of just two points $\vC_L$ and $\vC_R$ (see Fig.~\ref{fig:fig3}). 
Choose a coordinate frame such that the lens lies on the plane $\bX=0$, and the
centre of the lens is at the origin. In addition, let the ``world'' lie to the
left of the lens (that is with points $\bX < 0$), and the camera lie to the
right of the lens. 
Let $\bW$ consisting of points $\vX=(\bX,\bY,\bZ)$ \footnote{All vectors will be
column vectors. } with $\bX <-f$, and let $\bW '$ be the 3D image of the virtual
world, created by the lens, lying to the right of the lens. In particular, if
$\vX' = (\bX',\bY',\bZ')\in \bW'$, then $\bX'>f$. Here, $f$ is the focal length
of the lens.
Consider a plane of sensors (called a focal plane) lying inside the camera.
Assume that this is parallel to the lens, so it is a plane defined by the
equation $\rm{X} = F$, with $F>0$. The symbol of $F$ is chosen to imply that
this is the distance of the focal plane from the lens, but it is not the same as
the focal length of the lens, which is $f$. 

\vspace*{0.03in}
\noindent
{\bf Observation. }
{\em
An image $\vI_L$ of the world $\bW$ formed from rays passing through a point
$\vC_L$ in the focal plane is the same as a pin-hole camera image of the virtual
world $\bW'$ with projection centre $\vC_L$ and the same focal plane. The same
statement is true for the image $\vI_R$~formed from rays passing through
$\vC_R$.
}
\vspace*{-0.09in}

\begin{proof} Refer to Fig.~\ref{fig:fig3}. Consider a world point $\vX$, mapped
by the lens to a virtual world point $\vX'$. This means that if $\vC$ is any
point on the lens, then the ray $\vX \vC$ is refracted by the lens to the ray
$\vC \vX'$. In particular, this is true for the point $\vC_L$ lying on the lens.
Let the point $\vY$ be the point where the ray $\vC_L\vX '$ (extended if
necessary) meets the focal plane. Then $\vY$ is the point where $\vX$ is imaged
in $\vI_L$. This is because a ray traverses the path $\vX \vC_L \vY$, passing
through $\vX'$ on the way.
On the other hand, looked at in terms of the pinhole camera with centre $\vC_L$,
the ray from $\vC_L$ to $\vX'$ meets the focal plane at $\vY$. This means that
$\vY$ is the image of the point $\vX'$ in the pinhole image taken from $\vC_L$.
\end{proof}
\vspace{-2.0mm}

Therefore, if the two regions were single points $\vC_L$ and $\vC_R$, then the
two images $\vI_L$ and $\vI_R$ are exactly a stereo pair of images of the
virtual world $\bW'$.
Although the two points $\vC_L$ and $\vC_R$ are necessarily very close together,
the virtual world is also very small and very close to the camera lens, so there
will be appreciable disparity.

\begin{figure}[t]
    \centering
    \includegraphics[width=0.45\textwidth]{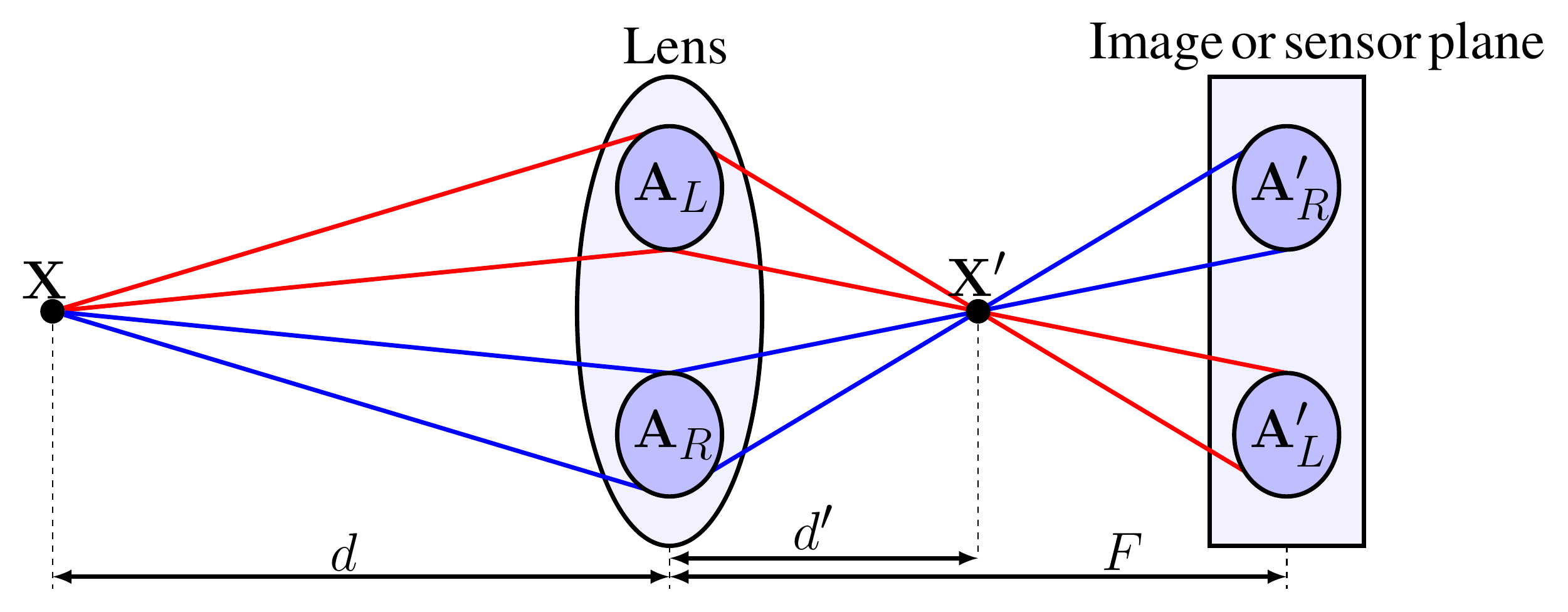}
    \vspace{-1mm}
    \caption{\it Rays from a point $\vX$ in the scene at depth $d$ pass through
the aperture $\vA_L$ and are focused at a point $\vX'$ at depth $d'$ in the
virtual world $\bW'$. The set of all the rays passing through $\vA_L$ form,
after refraction, a double-sided cone, with vertex at $\vX'$. This cone, meets
the focal plane at depth $F$ in a region that is geometrically similar to the
shape of $\vA_L$. This explanation is continued in Fig.~\ref{fig:fig5}.}
    \label{fig:fig4}
    \vspace{-2mm}
\end{figure}

\vspace{+0.5mm}
\noindent{\bf Non-pinhole model. }In the case where the two regions $\vA_L$ and
$\vA_R$ are bigger than a pinhole, and in fact constitute half the lens itself,
the images $\vI_L$ and $\vI_R$ formed will be made up as the superposition of
images of the virtual world $\bW'$ taken at all points $\vC_L$ in region $\vA_L$
and $\vC_R$ in region $\vA_R$. They will consequently be blurred. This is shown
in Fig.~\ref{fig:fig4}.

Therefore, the problem of finding depth in the scene from images $\vI_L$ and
$\vI_R$ is equivalent to doing stereo from blurred images. This will have its
problems. The blur will be depth dependent. Points in the virtual world that lie
on the focal plane $\pi$ will be in focus at independent of the position of the
point $\vC_L$ and $\vC_R$. Thus, points in the world corresponding to this
placement of the focal plane will be both in focus and identically positioned in
the two images $\vI_L$ and $\vI_R$. Points that lie off the focal plane will be
blurred and at the same time displaced by a disparity.

\subsection{Image synthesis from RGB-D image}\label{sec:rgbd}

Given an image with an associated depth map, it is possible to synthesize either
of the pair of dual-pixel images. Consider the image $\vI$, formed from all rays
passing through a region $\vA$ in the plane of the lens. (Note, $\vA$ is either
$\vA_L$ or $\vA_R$, and the corresponding image is either $\vI_L$ or $\vI_R$.)

Let $\vI_W$ be an RGB-D image of the world, taken nominally from the viewpoint
of the lens centre. Since the lens is normally very small with respect to the
world, it will be assumed that any world point visible from the lens centre will
be visible from any other point on the lens.

The RGB-D image gives us 3D coordinates of any point visible in the image. Let
$\vX = (\bX,\bY,\bZ)\in \bW$ be such a point. The image of this point in the
virtual world $\bW'$ is given by
\vspace*{-2mm}
\begin{equation}
    \vX' = (\bX',\bY',\bZ') = \frac{f}{f+\bX} (\bX,\bY,\bZ).
    \vspace*{-1mm}
\end{equation}
%
This point is then projected from each point $\vC\in \vA$ onto the focal plane
$\bX = F$. The projection rays, through $\vX'$ from points in $\vA$ form a cone
with cross-section $\vA$ and vertex $\vX'$. This cone meets the focal plane in a
further cross-section $\vA'$, Since the lens and focal plane are parallel, $\vA$
and $\vA'$ will be similar regions, in the sense that there is a similarity
transform relating $\vA$ and $\vA'$. Even more, this is simply a scaling and
translation (where the scale may be positive or negative). In particular, a
point $\vC \in \vA$ maps to a point $\vC' = s\vC +\vt$. Here, $\vt$ is a
2-dimensional offset and $s$ is a scale. For a given point $\vX$, the values of
$s$ and $\vt$ are constant, not dependent on the particular point $\vC$ chosen,
but they vary according to the point $\vX$ chosen.

\begin{figure}[t]
    \centering
    \includegraphics[width=0.45\textwidth]{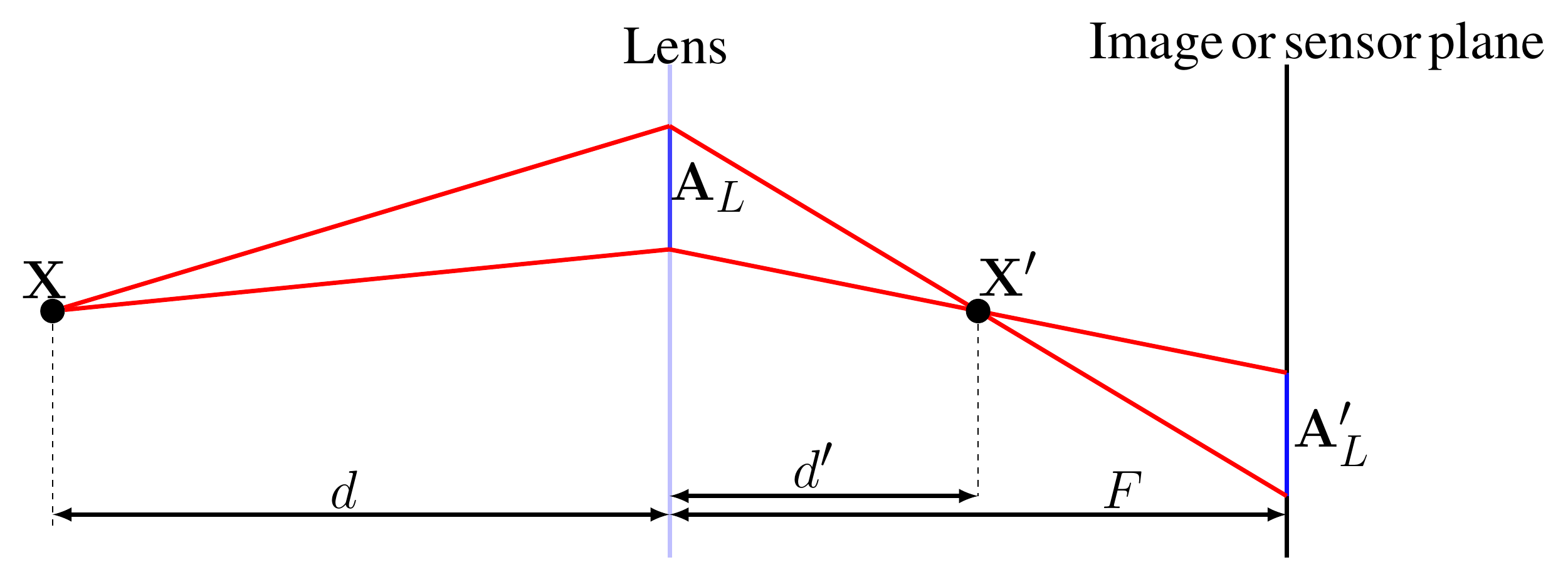}
    \vspace{-2mm}
    \caption{\it Viewed side-on, rays from a point $\vX$ pass through the
aperture $\vA_L$ (similarly $\vA_R$) and are focused at $\vX'$. They meet the
focal plane at depth $F$ (which may be either side of $\vX'$) covering a region
$\vA_L'$ geometrically similar to $\vA_L$. The scale of $\vA_L'$ is computed by
similar triangles; the scale factor is given by $ (d'-F)/d'$. This scale
factor is positive if $F$ lies in front of $\vX'$ and negative if it lies
behind; in this letter case, $\vA_L'$ is inverted with respect to $\vA_L$.  }
    \label{fig:fig5}
    \vspace{-2mm}
\end{figure}

Refer to Fig.~\ref{fig:fig5}. In particular, by similar triangles
%
\begin{equation}
    \vC' = T(\vC) = (1-s)\vC+s\vX' \ ,
\end{equation}
where $s = F/d'$.

\begin{figure*}[t]
\vspace*{-20pt}
    \centering
        \includegraphics[width=0.91\textwidth]{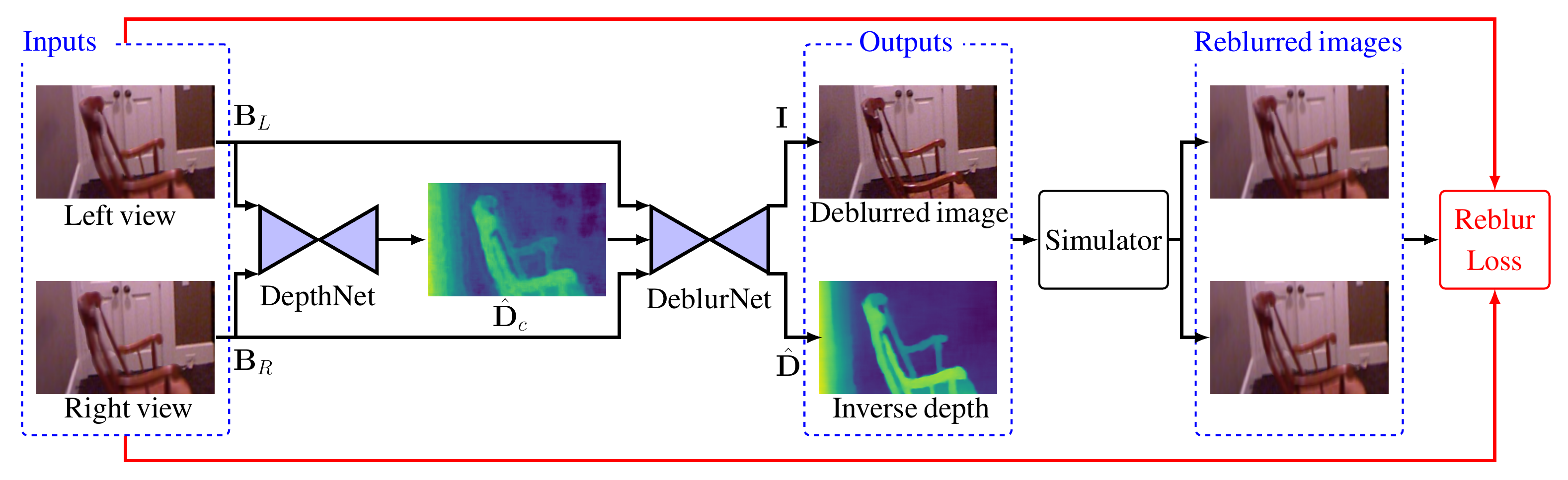}%
    \vspace{-2mm}
    \caption{\it The architecture of the proposed DDDNet. With the input DP
image pair, the DepthNet first estimates a coarse inverse depth map
$\hat{\vD}_c$. Then, we combine the coarse and blurred inverse depth map with
$\vB_{\{L,R\}}$, and feed them to the DeblurNet.
The deblurred image and estimated inverse depth  map are fed to our DP simulator to synthesize a DP image pair. The synthesized DP images are compared to the inputs using our reblur loss (Eq.~\eqref{eq:loss3}), to regularise the DDDNet in training. Note that, the ground truth
inverse depth maps and sharp images are used as supervision signals in training.
(Best viewed in colour on screen.)}
    \label{fig:pipeline}
    \vspace{-2mm}
\end{figure*}

\vspace{+0.2mm}
\noindent{\bf Image coordinates. } We start with an RGB-D image, so each pixel
in an image comes with an associated depth. For simplicity, we assume that all
distances (including depth) are measured in pixel coordinates.

We make the following assumptions. A pixel with coordinates \footnote{{It is
convenient to use $(y, z)$ for image coordinates, instead of the usual $(x,
y)$.}} $(y, z)$ corresponds with a ray in space defined by $-d(1,y/f,z/f)$ for
varying $d$. In particular, since each point in the image comes with a depth, we
are assuming that the point imaged at point $(y,z)$ has 3D coordinates
$(-d,-dy/f,-dz/f)$. This corresponds to a pinhole camera with focal plane given
by $\bX = f$ and camera centre (pinhole) at the origin.

A point $\vX = -d(1,y/f,z/f)$ at depth $d$ will be mapped to a point $\vX' =
d'(1,y/f,z/f)$, where $1/d+1/d'=1/f$, or
%
\begin{equation}
    d' = fd/(d-f) \ .
\end{equation}
Let $\vC = (0,\bY_0,\bZ_0)$ be a point in $\vA$ (lying on the lens plane $\bX =
0$). The line from $\vC$ through $\vX' = d'(1,y/f,z/f)$ is expressed as
$(1-s)\vC+s\vX'$ for varying values of $s$. This line meets the plane $\bX = F$
when $s = F/d'$. The coordinates of this point are therefore
%
\begin{equation}
\begin{split}
T(\vX, \vC) &= \frac{d'-F}{d'} \vC + \frac{F}{d'} \vX'\\
&= \frac{d'-F}{d'} \vC + F (1,y/f,z/f) \ .
\end{split}
\end{equation}
We are only interested in the $(\bY,\bZ)$-coordinates, in which case we have
%
\begin{equation}\label{eq:TFunc}
    T(\vX, \vC) =\frac{d'-F}{d'} (\bY_0,\bZ_0) + F (y/f,z/f) \ .
\end{equation}
 
Here, the transformation function $T(\vX, \vC)$ for the left and right image has
been defined with points $\vC_L$ and $\vC_R$ in $\vA_L$ and $\vA_R$
respectively. 

\subsection{DP simulator}\label{sec:simu}
Given a sharp image $\vL\in \R^{H\times W}$ and its associated depth map $\vd\in
\R^{H\times W}$, we can simulate a DP image pair $\vB_{\{L,R\}}$ as follows.
Here, $H$ and $W$ are the image height and width, and
it is also assumed that focal length $f$ and focal-plane-sensor distance $F$ are known
(measured in units of pixel-size).

For each pixel $(y, z)$ in $\vI$ , as shown in Fig.~\ref{fig:fig5} the
intensity of a pixel is spread over regions $\vR_L$ and $\vR_R$, in the
left and right  view of the DP pair. 
Each region contains a set of points $\vp$, containing $|\vR|$ pixels,
and the intensity $\vI(y, z)$ of the pixel $(y, z)$ is spread out evenly over this set of pixels.
Now running over all pixels $\vI(y, z)$, and summing, a pair of 
dual-pixel images are created.  This is essentially blurring with
depth-dependent blur kernel.

This operation is expensive computationally, since it requires looping
over each pixel $(y, z)$ in the image, as well as each pixel in the
region $\vR$, thus $4$ levels of looping.

We utilize the concept of `integral image' to speed up 
the computation, so that it has complexity independent
of the size of the region $\vR$, thus of order $\mathcal{O}(n)$ where $n$ is the number of
pixels.
In this description, we assume that the left
and right halves of the aperture are approximated by rectangles, though with some
amount of extra computation, more general shapes may be accommodated.

Given a pixel $(y,z)$ of the RGB-D image at a given depth, its corresponding light rays
will pass through every point of $\vA_L$. To compute the blur-region $\vR_L$,
one needs only to compute the destinations of the rays passing through
the four corners of $\vA_L$, which will be denoted by
$\vp_{tl}= (y_{tl}, z_{tl})$, $\vp_{tr}= (y_{tr},z_{tr})$, $\vp_{bl}=
(y_{bl}, z_{bl})$ and $\vp_{br}= (y_{br}, z_{br})$ in the left image $\vB_L$.
The positions of these four points are given by Eq.~\eqref{eq:TFunc}. Here, the
subscript $_{tl}$, $_{tr}$, $_{bl}$ and $_{br}$ denotes top-left, top-right, bottom-left and
bottom-right, respectively. 

We create a {\em differential mask} $\calI_L \in \R^{H\times W}$ 
for the region $\vR_L$ defined by
\begin{equation}
\begin{aligned}
 \calI_L(\vp_{tl})& =  \vI(y,z)/|\vR_L|, 
&\calI_L(\vp_{tr})& =  - \vI(y,z)/|\vR_L|,\\ 
\calI_L(\vp_{bl})& = - \vI(y,z)/|\vR_L|,
&\calI_L(\vp_{br})& =  \vI(y,z)/|\vR_L| ~ .   
\end{aligned}
\end{equation}
These values are summed for the regions $\vR_L$ corresponding to all points
$(y, z)$ in the image, to create the {\em differential image}. 
Finally, we integrate the differential image, and the left/right view of the DP
image pair is given by
%
\begin{equation}\label{eq:}
\hat{\vB}_{\{L,R\}} = \calT(\calI_{\{L,R\}}),
\end{equation}
where $\calT(\cdot)$ denotes the integral process. 
The integration process~\cite{viola2001robust} is also closely related to `summed area tables' in
graphics~\cite{crow1984summed}. 

Our simulator allows the vision community to
collect large amounts of DP data with ground truth, opening the door to
accurately benchmark DP-based methods. Our simulator can also be used to
supervise the learning process. 

With the DP model and the simulator, we build our DDDNet considering both stereo
and defocus cues.

\section{DP-based Depth and Deblur Network}\label{sec:DDDnet}
\subsection{Network architecture.} 

The input of our DDDNet is the left and right view of a DP image pair $\vB_{\{L,R\}}$. The output of our DDDNet is the estimated inverse depth map
$\hat{\vD}$ and the deblurred image $\hat{\vL}$. We use ground-truth latent
sharp image $\vL$ and inverse depth map $\vD$ for training.

The pipeline of DDDNet is shown in Fig.~\ref{fig:pipeline}. It consists of two
components: DepthNet
$g(\cdot)$ with parameters $\vG$ and DeblurNet $f(\cdot)$ with parameters $\vF$. The DepthNet is based
on~\cite{zhong2020nipsstereo} and the DeblurNet is based on the multi-patch
network~\cite{Zhang_2019_CVPR}. 
Note that our approach is independent to the choices of $f(\cdot)$ and
$g(\cdot)$ (\eg, multi-scale and multi-patch architectures). We start with a
coarse inverse depth map $\hat{\vD}_c$ estimation by the DepthNet, where
$\hat{\vD}_c = g(\vB_{\{L,R\}};\vG)$. 
Then, we combine the coarse and blurred inverse depth map with $\vB_{\{L,R\}}$, and
feed it to the DeblurNet. The DeblurNet is an encoder-decoder network, and
$\{\hat{\vL},\hat{\vD}\} =f(\vB_{\{L,R\}},\hat{\vD}_c ;\vF)$.

\subsection{Loss functions.} 
We use a combination of an image restoration loss, depth loss, and an image
reblur loss. The final loss is a sum of the three losses.
\vspace{-2mm}
\begin{equation} \label{eq:loss}
\begin{aligned}
\calL = \calL_{\rm res} + \calL_{\rm d} + \calL_{\rm reb} \ .
\end{aligned}
\end{equation}

{\noindent{\bf Image restoration loss $\calL_{\rm res}$.}} 
This loss ensures the deblurred image is similar to the target image, and is given 
by
\vspace{-1mm}
\begin{equation} \label{eq:loss1}
\begin{aligned}
\calL_{\rm res} = \frac{1}{N}\sum_{y,z}\|\vL(y,z)-\hat{\vL}(y,z) \| \ ,
\end{aligned}
\end{equation}
where $N$ is the number of pixels and $\|\cdot\|$ is the $\ell_2$ norm.

\vspace{+1mm}
{\noindent{\bf Depth loss $\calL_{\rm d}$.}} 
We adopt the smooth $\ell_1$ loss. 
Smooth $\ell_1$ loss $\calS(\cdot)$ is widely used in a regression task, for its robustness and low sensitivity to
outliers~\cite{girshick2015fast}. The $\calL_{\rm d}$ is defined as
\vspace{-2mm}
\begin{equation} \label{eq:loss2}
\begin{aligned}
\calL_{\rm d} &= \frac{1}{N}\sum_{y,z} \calS(\vD(y,z)-\hat{\vD}(y,z)) \ . \\
\end{aligned}
\end{equation}

{\noindent{\bf Image reblur loss $\calL_{\rm reb}$.}} 
The reblur loss penalizes the differences between the input blurred images and
the reblurred images (from the DP simulator).
The reblur loss explicitly enforces the restored image and inverse depth map to lie on the manifold of the ground-truth image and inverse depth map, subject to our DP model. The $\calL_{\rm reb}$ is given by 

\vspace{-2mm}
\begin{equation} \label{eq:loss3}
\begin{aligned}
\calL_{\rm reb}=
\frac{1}{N}\sum_{y,z}\|\vB_{\{L,R\}}(y,z)-\hat{\vB}_{\{L,R\}}(y,z)\| \  .
%
\end{aligned}
\end{equation}
%
\begin{figure}[t]
\begin{center}
\begin{tabular}{ccc}
\hspace{-0.25 cm}
\includegraphics[width=0.1541\textwidth,height=0.1031
\textwidth]{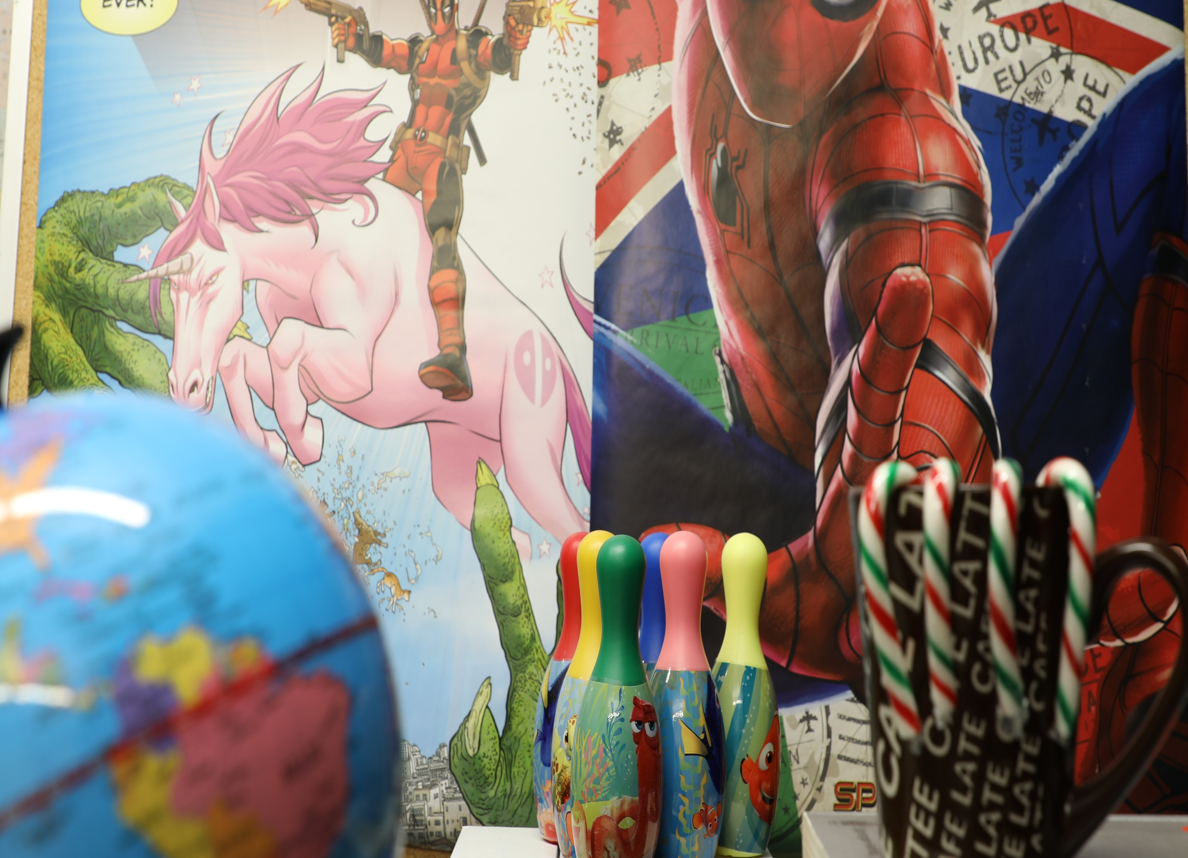}  
\hspace{-0.45 cm}
&\includegraphics[width=0.1541\textwidth,height=0.1031
\textwidth]{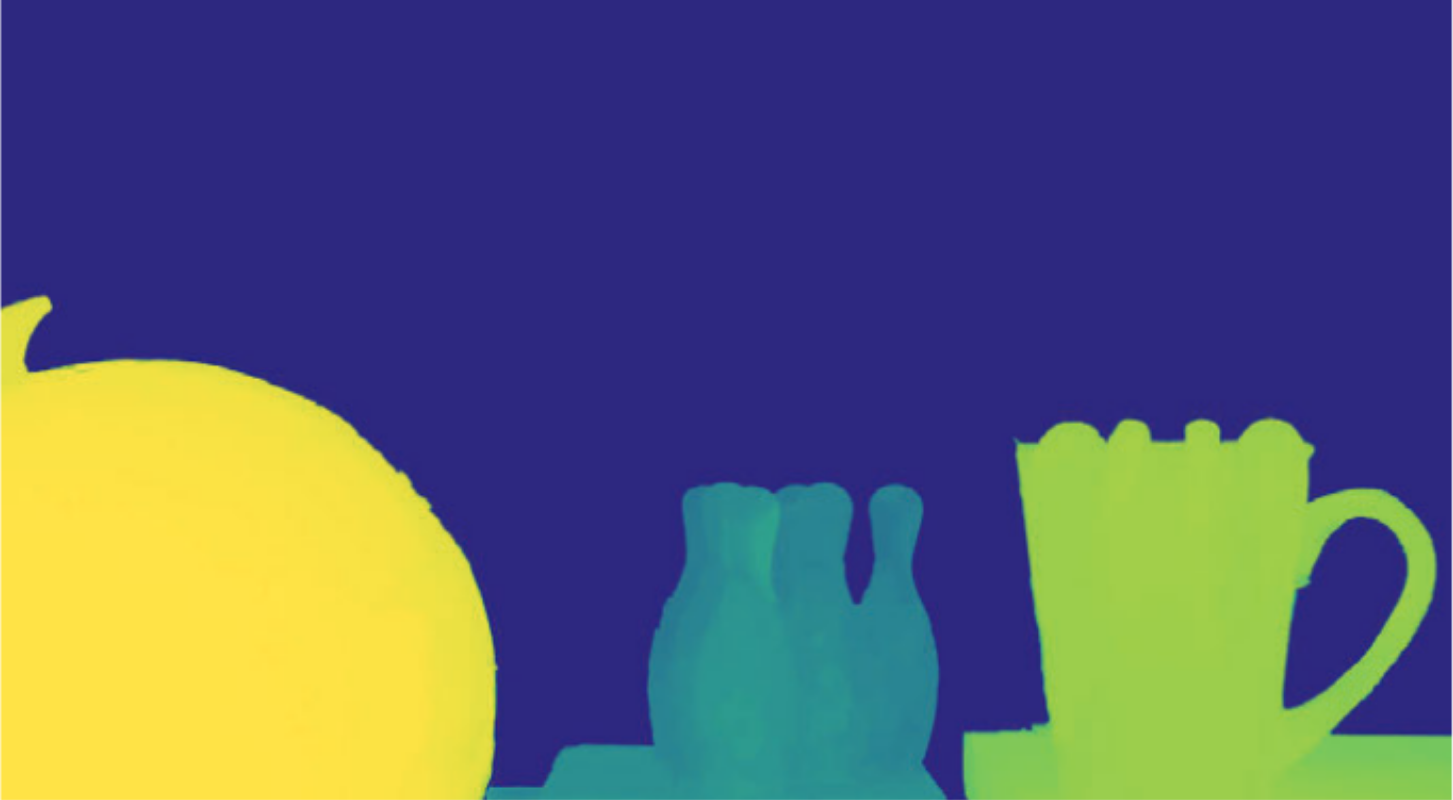}   
\hspace{-0.45 cm}
&\includegraphics[width=0.1541\textwidth,height=0.1031
\textwidth]{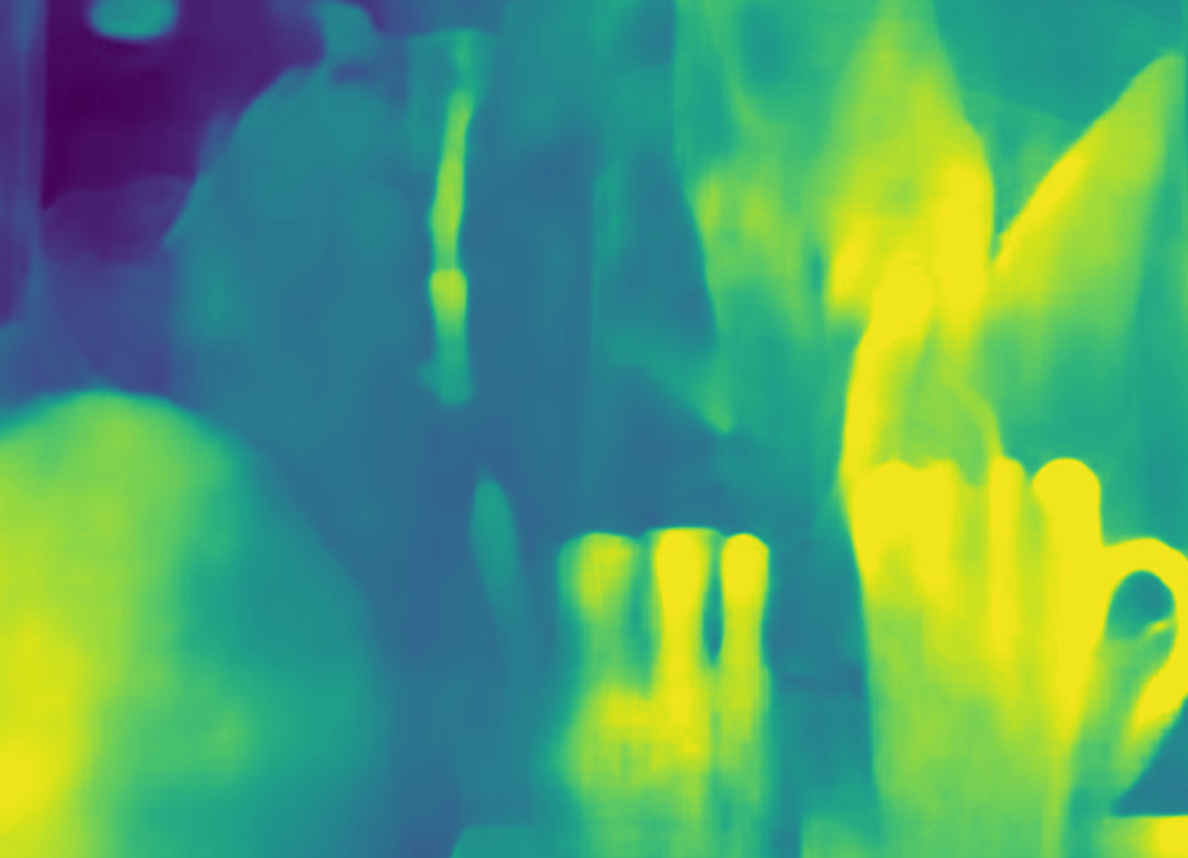}\\
\hspace{-0.25 cm}
(a) Input image
\hspace{-0.45 cm}
&(b) GT:  Depth
\hspace{-0.45 cm}
&(c) BTS \cite{lee2019big}\\
\hspace{-0.25 cm}
\includegraphics[width=0.1541\textwidth,height=0.1031
\textwidth]{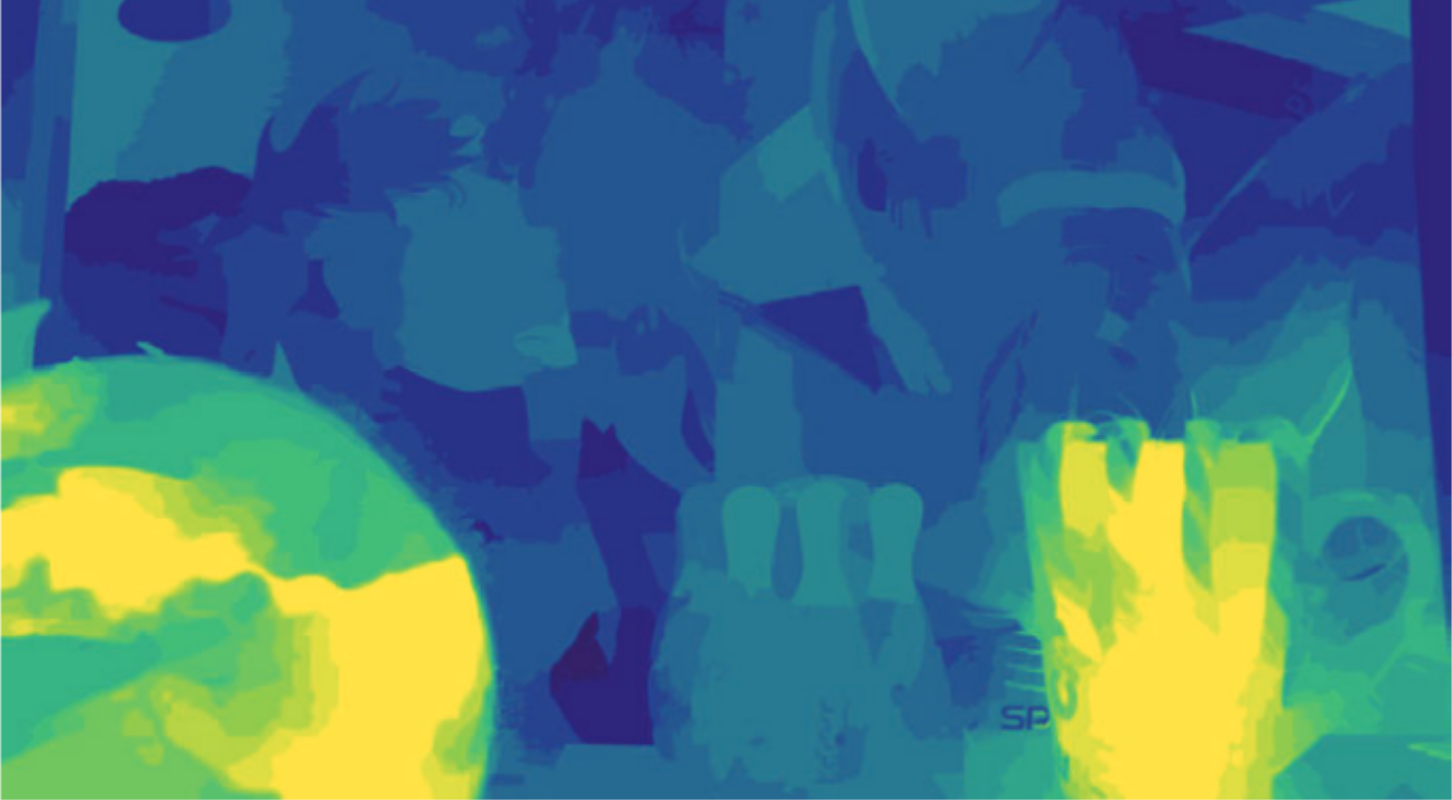}  
\hspace{-0.45 cm}
&\includegraphics[width=0.1541\textwidth,height=0.1031
\textwidth]{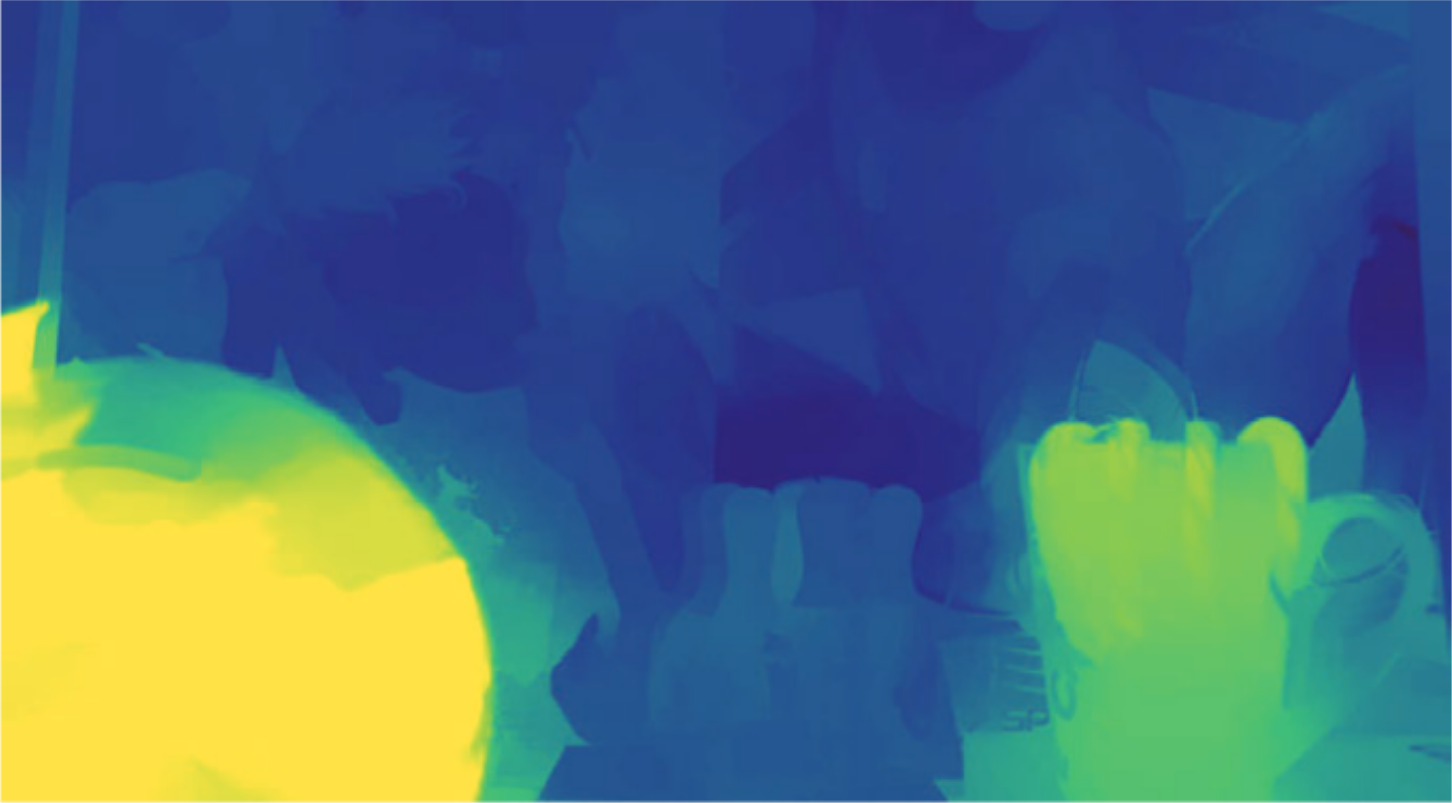}  
\hspace{-0.45 cm}
&\includegraphics[width=0.1541\textwidth,height=0.1031
\textwidth]{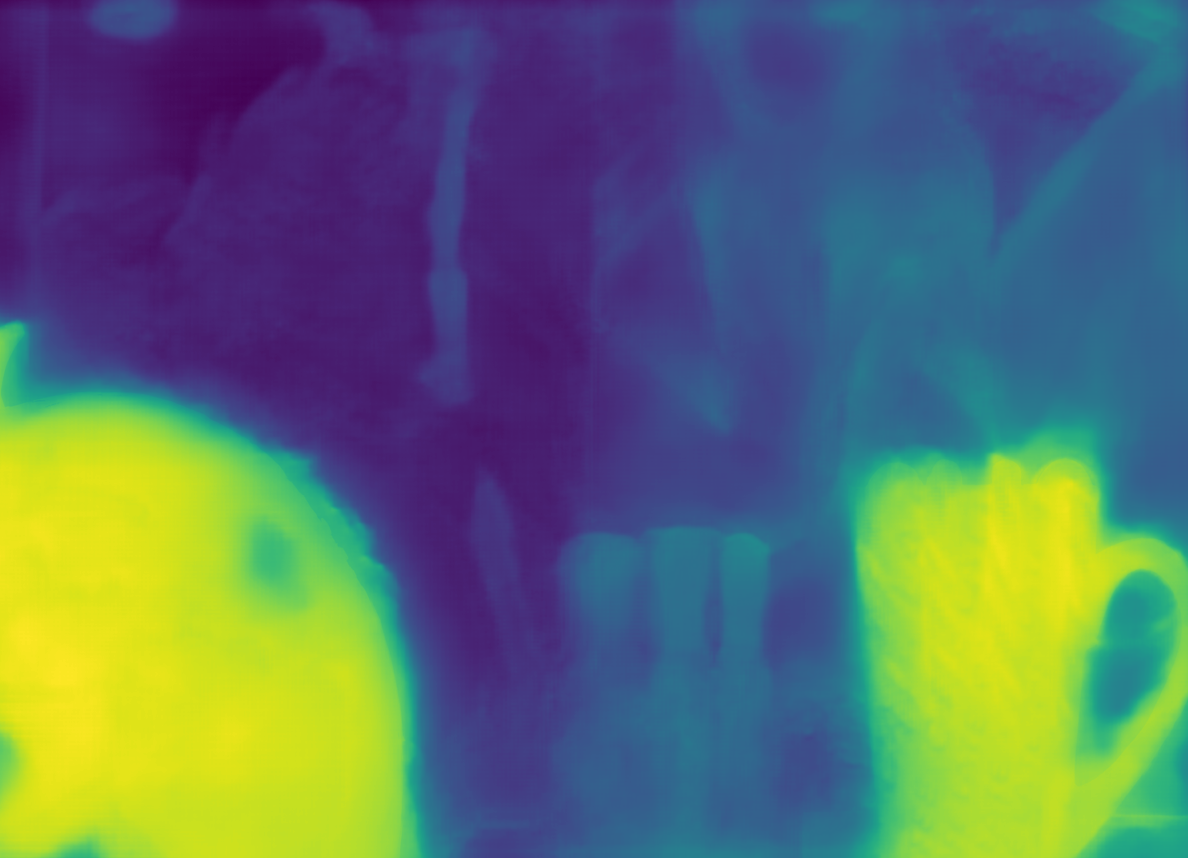}\\   
\hspace{-0.25 cm}
(d) SDoF \cite{wadhwa2018synthetic}
\hspace{-0.45 cm}
&(e) DPdisp \cite{punnappurath2020modeling}
\hspace{-0.45 cm}
&(f) Ours\\
\end{tabular}
\vspace{-3mm}
\end{center}
\caption{\label{fig:dpd-disp} \it 
An example of inverse depth estimation result on the real DPD-disp
dataset~\cite{punnappurath2020modeling}. (Best viewed in colour on screen.)
}
\vspace{-1mm}
\end{figure}

\begin{table}[]
\caption{ \label{tab:Deblur-table}
\em Quantitative analysis of deblurring results on the DPD-blur dataset
\cite{abuolaim2020defocus}, Our-syn dataset, and Our-real dataset. The best
results are shown in bold. 
Here, `Ours$_{\rm wb}$' and `Ours$_{\rm reb}$' denotes our model trained without
and with the reblur loss, respectively.
The results demonstrate that our model improves significantly with the reblur
loss. Besides `PSNR' (in dB), to explicitly illustrate the improved quality of
deblurred images, we use a metric `RMSE\_rel' (RMSE$/255$, in $\%$) to further
explain `what $1$dB improvement means'. Compared with the second-best method,
the relative improvement of our method in intensity value for the three datasets
is $13\%$, $19\%$, and $12\%$, respectively.
}
 \aboverulesep=0ex
 \belowrulesep=0ex
\begin{adjustbox}{width=0.475\textwidth}
\begin{tabular}{ccccc}
\toprule
\multicolumn{5}{c}{DPD-blur}                               \\ \bottomrule
\multicolumn{1}{c}{}     & \multicolumn{1}{c}{EBDB \cite{karaali2017edge}}  &
\multicolumn{1}{c}{DMENet \cite{lee2019deep}}    & \multicolumn{1}{c}{DPDNet
\cite{abuolaim2020defocus}} & Ours$_{\rm wb}$/Ours$_{\rm reb}$  \\ \hline
\multicolumn{1}{l}{PSNR $\uparrow$} & \multicolumn{1}{c}{24.82} &
\multicolumn{1}{c}{23.93} & \multicolumn{1}{c}{25.53}  & {\bf 26.15/26.76} \\
\hline
\multicolumn{1}{l}{SSIM $\uparrow$} & \multicolumn{1}{c}{0.801} &
\multicolumn{1}{c}{0.812} & \multicolumn{1}{c}{0.826}  & {\bf 0.827/0.842}      
\\ \hline
\multicolumn{1}{l}{MSE\_rel $\downarrow$} & \multicolumn{1}{c}{5.74} &
\multicolumn{1}{c}{6.36} & \multicolumn{1}{c}{5.29}  & {\bf 4.93/4.59} \\
\toprule
\multicolumn{5}{c}{Our-syn}                                      \\ \bottomrule
\multicolumn{1}{l}{PSNR $\uparrow$} & \multicolumn{1}{c}{26.48}      &
\multicolumn{1}{c}{30.14}        &  \multicolumn{1}{c}{31.45}       & {\bf
32.17/33.21 }      \\ \hline
\multicolumn{1}{l}{SSIM $\uparrow$} & \multicolumn{1}{c}{0.891} &
\multicolumn{1}{c}{0.939} & \multicolumn{1}{c}{0.926} & {\bf 0.948/0.956}  \\
\hline
\multicolumn{1}{l}{MSE\_rel $\downarrow$} & \multicolumn{1}{c}{4.74} &
\multicolumn{1}{c}{3.11} & \multicolumn{1}{c}{2.68}  & {\bf 2.46/2.17} \\
\toprule
\multicolumn{5}{c}{Our-real}                                      \\
\bottomrule
 \multicolumn{1}{l}{PSNR $\uparrow$} & \multicolumn{1}{c}{21.67} &
\multicolumn{1}{c}{23.18} & \multicolumn{1}{c}{22.65} & \bf{23.99/24.03} \\
\hline
\multicolumn{1}{l}{SSIM $\uparrow$} & \multicolumn{1}{c}{0.763} &
\multicolumn{1}{c}{0.809} & \multicolumn{1}{c}{0.808} & \bf 0.826/ \bf 0.850  \\
\hline
\multicolumn{1}{l}{MSE\_rel $\downarrow$} & \multicolumn{1}{c}{8.25} &
\multicolumn{1}{c}{6.39} & \multicolumn{1}{c}{7.09}  & {\bf 6.24/ \bf 6.13} \\
\bottomrule
\end{tabular}
\end{adjustbox}
\end{table}

\begin{table}[]
\caption{ \label{tab:DPD-disp}
\em Quantitative analysis of depth estimation on the DPD-disp
dataset~\cite{punnappurath2020modeling}. 
As the DPD-disp dataset only provides testing data, we directly use our model
(trained on Our-syn dataset, namely `Ours') without fine-tuning and achieves the
second best. Then, we use our reblur loss to fine-tuning our model (without
using the GT depth map). Our model with the reblur loss `Ours$_{\rm ft}$
(fine-tune)' achieves competitive results, indicating that our simulated DP
outputs resemble those from a real DP sensor.
}
 \aboverulesep=0ex
 \belowrulesep=0ex
\begin{adjustbox}{width=0.475\textwidth}
\begin{tabular}{lcccc}
\toprule
          & AI(1) $\downarrow$ & AI(2) $\downarrow$ & 1-$\rho_s$ $\downarrow$&
\begin{tabular}[c]{@{}c@{}}Geometric  \\ Mean\end{tabular}$\downarrow$ \\
\hline
BTS \cite{lee2019big} & 0.1070   & 0.1767    & 0.6149     & 0.2686    \\ \hline
Monodepth2 \cite{monodepth2} & 0.1139  & 0.1788 &0.6153 &0.2285     \\ \hline 
SDoF~\cite{wadhwa2018synthetic} & 0.0875   &  0.1294   & 0.2910             & 
0.1443    \\ \hline 
DPdisp \cite{punnappurath2020modeling} & {\bf 0.0481 }   & {\bf  0.0845}  
&0.1037             &  {\bf 0.0671}    \\ \hline
Ours   & 0.0906    & 0.1291    & 0.2456    & 0.1207  \\ \hline   
`Ours$_{\rm ft}$'  &0.0609    &0.0985    &{\bf 0.1026}    & 0.1098
            \\ \bottomrule
\end{tabular}
\end{adjustbox}
\end{table}

\begin{table*}[]
\caption{\label{tab:syn}
\em Quantitative analysis of deblurring and depth estimation results on our
synthetic dataset. The synthetic dataset is generated using our DP simulator.
Here, we evaluate the intermediate outputs of DDDNet (see
Fig.~\ref{fig:pipeline}) step by step, `Ours$_{\rm b}$' denotes the coarse depth
map $\hat{\vD}_c$ from DepthNet, `Ours$_{\rm wb}$' denotes the depth map
$\hat{\vD}$ from DeblurNet, and `Ours$_{\rm reb}$' denotes the final results
with the reblur loss. BTS achieves lower in `RMSE' but higher in `RMSE\_log',
and the reason is that BTS gets a number of zeros in their results.   
}
 \aboverulesep=0ex
 \belowrulesep=0ex
\begin{adjustbox}{width=0.99\textwidth}
\begin{tabular}{llcccccccc}
\toprule
Sensor    &Method & abs\_rel $\downarrow$ & sq\_rel $\downarrow$ & rmse
$\downarrow$ & rmse\_log $\downarrow$ & $a<1.25$ $\uparrow$     & $a<1.25^2$
$\uparrow$    & $a<1.25^3$ $\uparrow$ & PSNR $\uparrow$    \\ \hline
Monocular & BTS~\cite{lee2019big}  & 0.377     & 0.195   & {\bf 0.255}   & 0.484
  & 0.554 & 0.741 & 0.994 & -\\ \hline
Stereo & AnyNet~\cite{wang2019anytime}  & 0.128     & 0.117   & 0.731 & 0.168   
 & 0.828 & 0.923 & 0.998 & - \\ \hline
\multirow{4}{*}{Dual-pixel}
& DPdisp~\cite{punnappurath2020modeling} & 0.328  & 0.479   & 1.252 & 0.332    &
0.438  & 0.804 & 0.965 & - \\ \cline{2-10}
& Ours$_{\rm b}$ &  0.149  &   0.364  &   1.222  & 0.224  &   0.743  & 0.930  &
0.978 & - \\ \cline{2-10}
& Ours$_{\rm wb}$ &  0.091  &   0.082  &   0.599  &  0.123  &   0.918  & 0.993 
& 0.998 & 32.171 \\ \cline{2-10}
& Ours$_{\rm reb}$ & {\bf 0.083}  & {\bf 0.052}  & 0.461  &  {\bf 0.111}  & {\bf
0.936}  & {\bf 0.998}  & {\bf 1.000} & {\bf 33.218} \\ \bottomrule
\end{tabular}
\end{adjustbox}
\end{table*}
\vspace{-3mm}
\section{Experiment}
\subsection{Experimental setup}
\noindent{\bf{Real dataset.}}
We evaluate our method on three real DP datasets, namely, the Defocus Depth
estimation dataset (DPD-disp)~\cite{punnappurath2020modeling}, Defocus Deblur
Dual-Pixel dataset (DPD-blur)~\cite{abuolaim2020defocus}, and our new-collected
dataset (Our-real) . 
DPD-disp dataset provides DP images with depth maps. The ground-truth (GT) depth
map is computed by applying the well-established depth-from-defocus technique.
DPD-blur provides a collection of image pairs captured by Canon 5D IV with
all-in-focus and out-of-focus.  
Compared with the DPD-blur dataset, Our-real is collected by Canon with a
variety of aperture sizes, varying from $f/4$ to $f/22$. Each all-in-focus image
is associated with several out-of-focus blurred images, yielding a diversity
dataset. 
Our-real contains 150 scenes, including both indoor and outdoor scenes, captured
under a variety of lighting conditions.

\vspace*{+0.2 mm}
\noindent{\bf{Synthetic dataset.}}
Our synthetic dataset is generated using the proposed DP simulator. We use the
NYU depth dataset~\cite{SilbermanECCV12} as the input to our simulator, as it
provides RGB images with depths. 
Giving different camera parameters, we simulate 5000 image pairs for training
and 500 image pairs for testing.  


\vspace*{+0.2 mm}
\noindent{\bf Evaluation metrics.} 
Given a DP-pair, our method jointly estimates a depth map and a deblurred image.
We use standard metrics to evaluate the quality of estimated depth map and
restored image separately. For depth map, we use absolute relative error
`Abs\_Rel', square relative error `Sq\_Rel', root mean square error `RMSE' and
its log scale `RMSE\_log', and the $\delta$ inlier ratios (maximal mean relative
error of $\delta_i = 1.25^i$ for $i\in {1,2,3}$). For restored image, we adopt
the peak signal-to-noise ratio `PSNR', structural similarity `SSIM', and RMSE
relative error `RMSE\_rel'. For the DPD-disp dataset, we follow
DPdisp~\cite{punnappurath2020modeling} to use the affine invariant version of
MAE ( `AI(1)'), RMSE (`AI(w)'), and Spearman's rank correlation (`$1-|\rho_s|$')
for evaluation.

\vspace*{+0.2 mm}
\noindent{\bf Baseline methods.} 
For depth map, we compare with state-of-the-art monocular
(BTS~\cite{lee2019big}, Monodepth2~\cite{monodepth2}), stereo
(AnyNet~\cite{wang2019anytime}), and DP (SDoF~\cite{wadhwa2018synthetic}, 
DPdisp~\cite{punnappurath2020modeling}) based depth estimation methods.
For defocus deblurring, we compared with EBDB~\cite{karaali2017edge},
DMENet~\cite{lee2019deep}, and DPDNet~\cite{abuolaim2020defocus}. All methods
are evaluated on each dataset independently. All learning methods are fine-tuned
for each dataset (except DPD-disp, as no training data is provided). 
To fine-tune DMENet~\cite{lee2019deep} on the DPD-blur and Our-real dataset, we
use BTS~\cite{lee2019big} to estimate a coarse depth map to train, as the two
datasets do not provide ground-truth depth maps. The same coarse depth maps are
also used to train our network.

\vspace*{+0.2 mm}
\noindent{\bf{Implementation details.}} 
Our network is implemented in Pytorch and is trained from scratch using the Adam
optimizer~\cite{kingma2014adam} with a learning rate of $10^{-4}$ and a batch
size of $10$. Our model is trained on a single NVIDIA Titan XP GPU. The code and
data will be released. 


\begin{figure*}[ht]
\begin{center}
\resizebox{0.94\textwidth}{!}{
\begin{tabular}{ccccc}
\hspace{-0.25 cm}
\includegraphics[width=0.227\textwidth,height=0.1571
\textwidth]{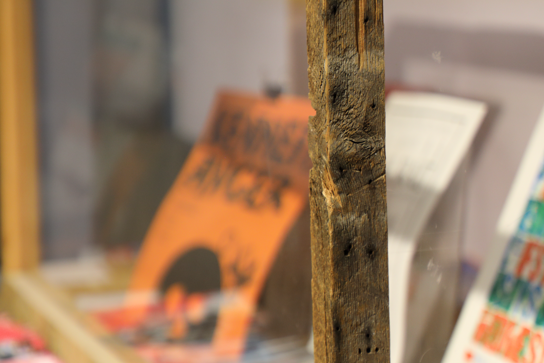}  
\hspace{-0.45 cm}
&\includegraphics[width=0.227\textwidth,height=0.1571
\textwidth]{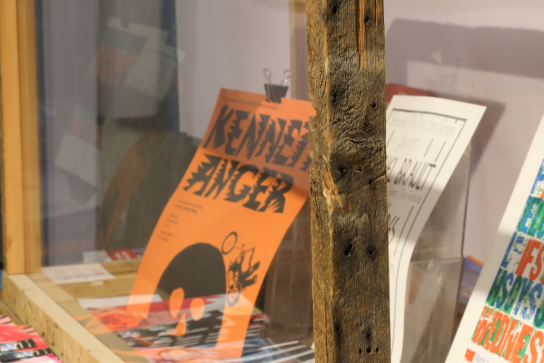} 
\hspace{-0.45 cm}
&\includegraphics[width=0.257\textwidth,height=0.1571
\textwidth]{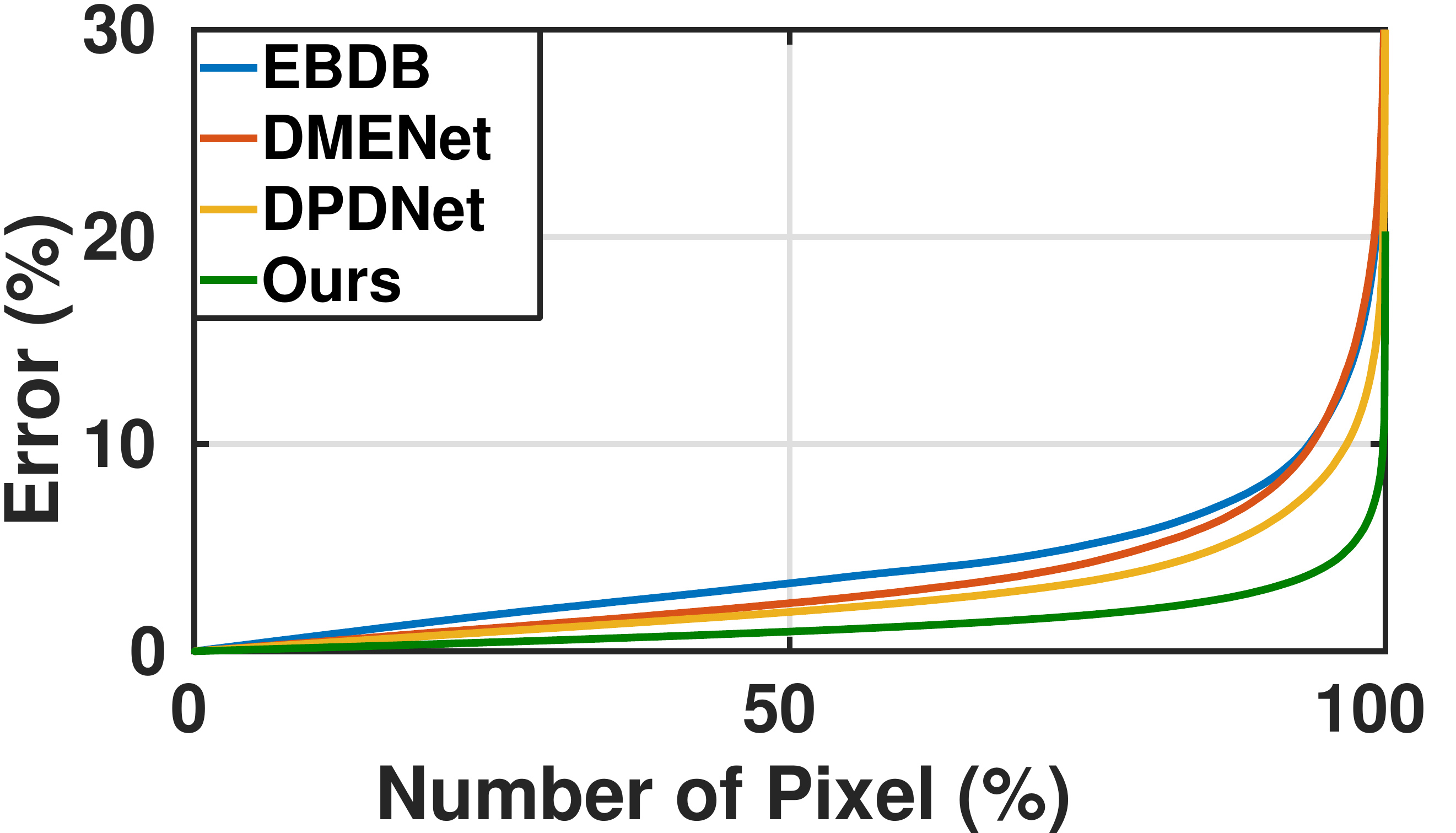} 
\hspace{-0.45 cm}
&\includegraphics[width=0.227\textwidth,height=0.1571
\textwidth]{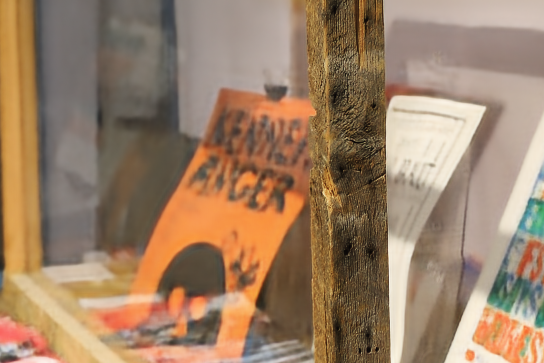} 
\hspace{-0.45 cm}
&\includegraphics[width=0.227\textwidth,height=0.1571
\textwidth]{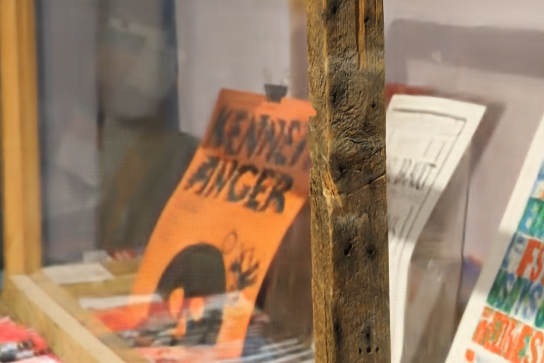} \\
\hspace{-0.25 cm}
\includegraphics[width=0.227\textwidth,height=0.1571
\textwidth]{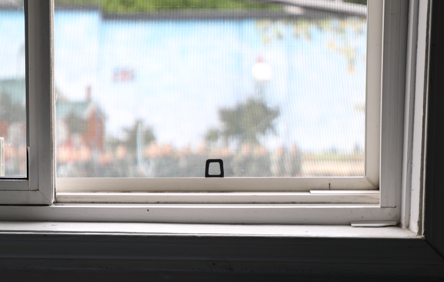}  
\hspace{-0.45 cm}
&\includegraphics[width=0.227\textwidth,height=0.1571
\textwidth]{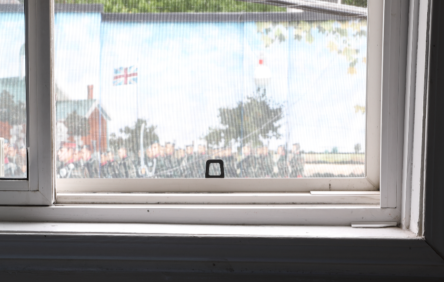} 
\hspace{-0.45 cm}
&\includegraphics[width=0.257\textwidth,height=0.1571
\textwidth]{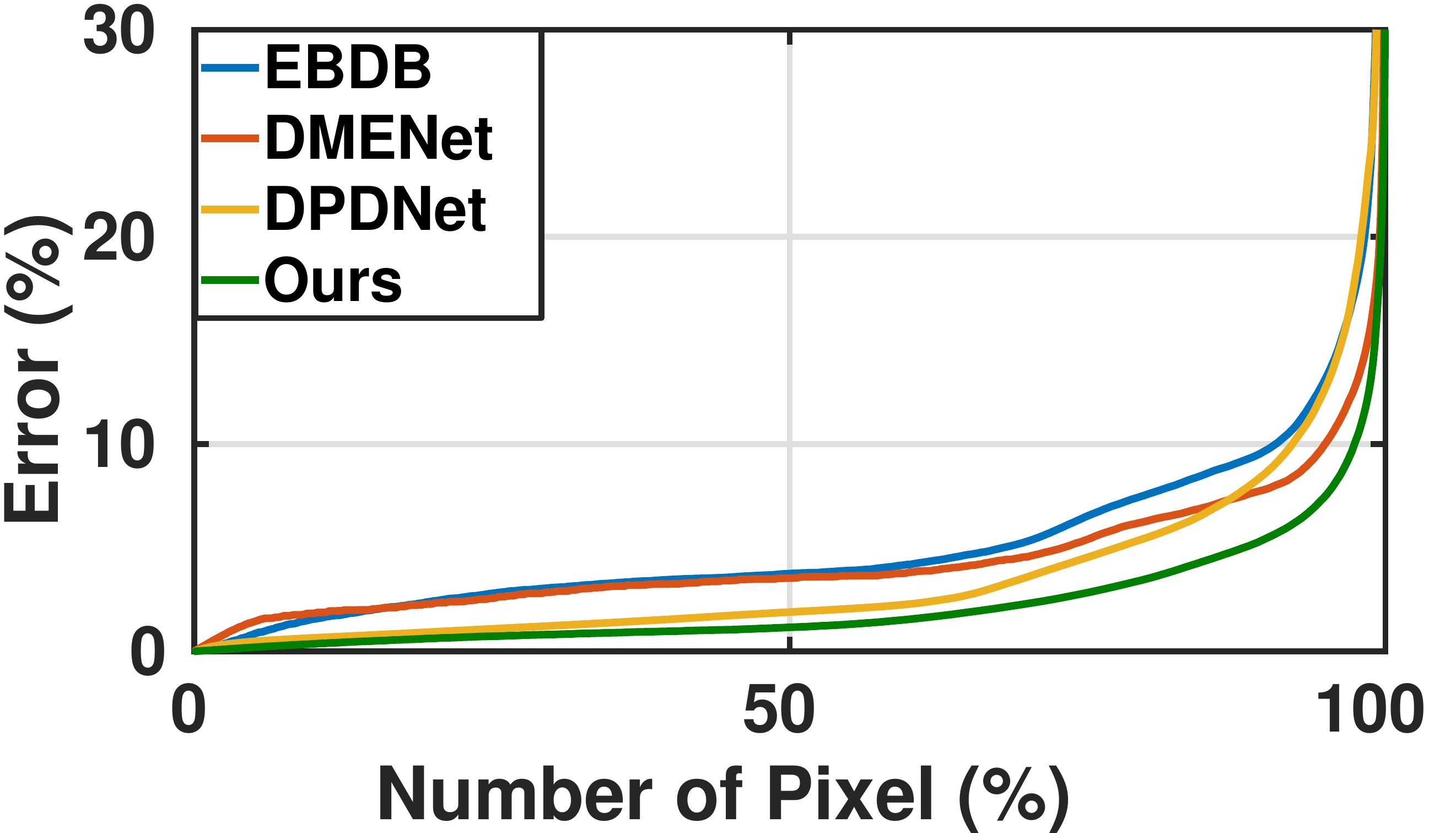} 
\hspace{-0.45 cm}
&\includegraphics[width=0.227\textwidth,height=0.1571
\textwidth]{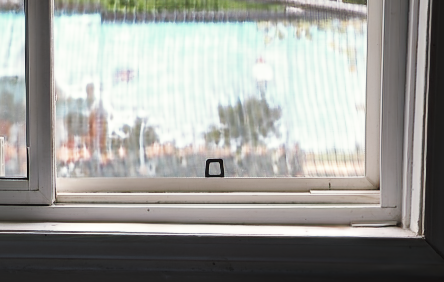} 
\hspace{-0.45 cm}
&\includegraphics[width=0.227\textwidth,height=0.1571
\textwidth]{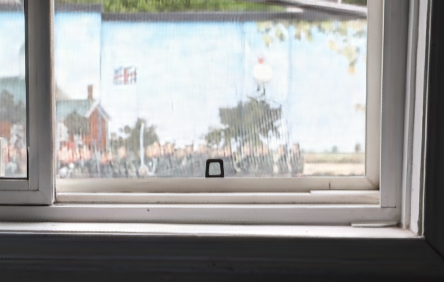}\\ 
\hspace{-0.25 cm}
\includegraphics[width=0.227\textwidth,height=0.1571
\textwidth]{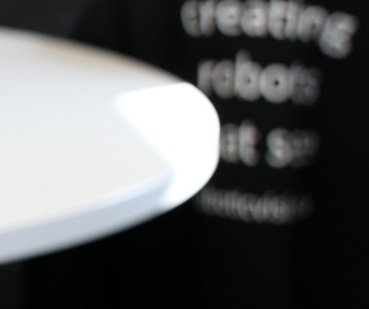}  
\hspace{-0.45 cm}
&\includegraphics[width=0.227\textwidth,height=0.1571
\textwidth]{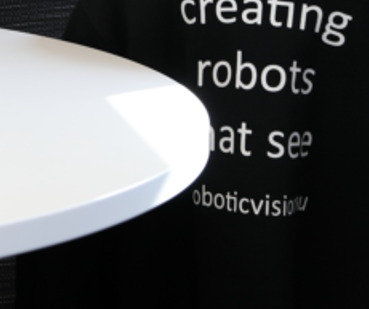} 
\hspace{-0.45 cm}
&\includegraphics[width=0.257\textwidth,height=0.1571
\textwidth]{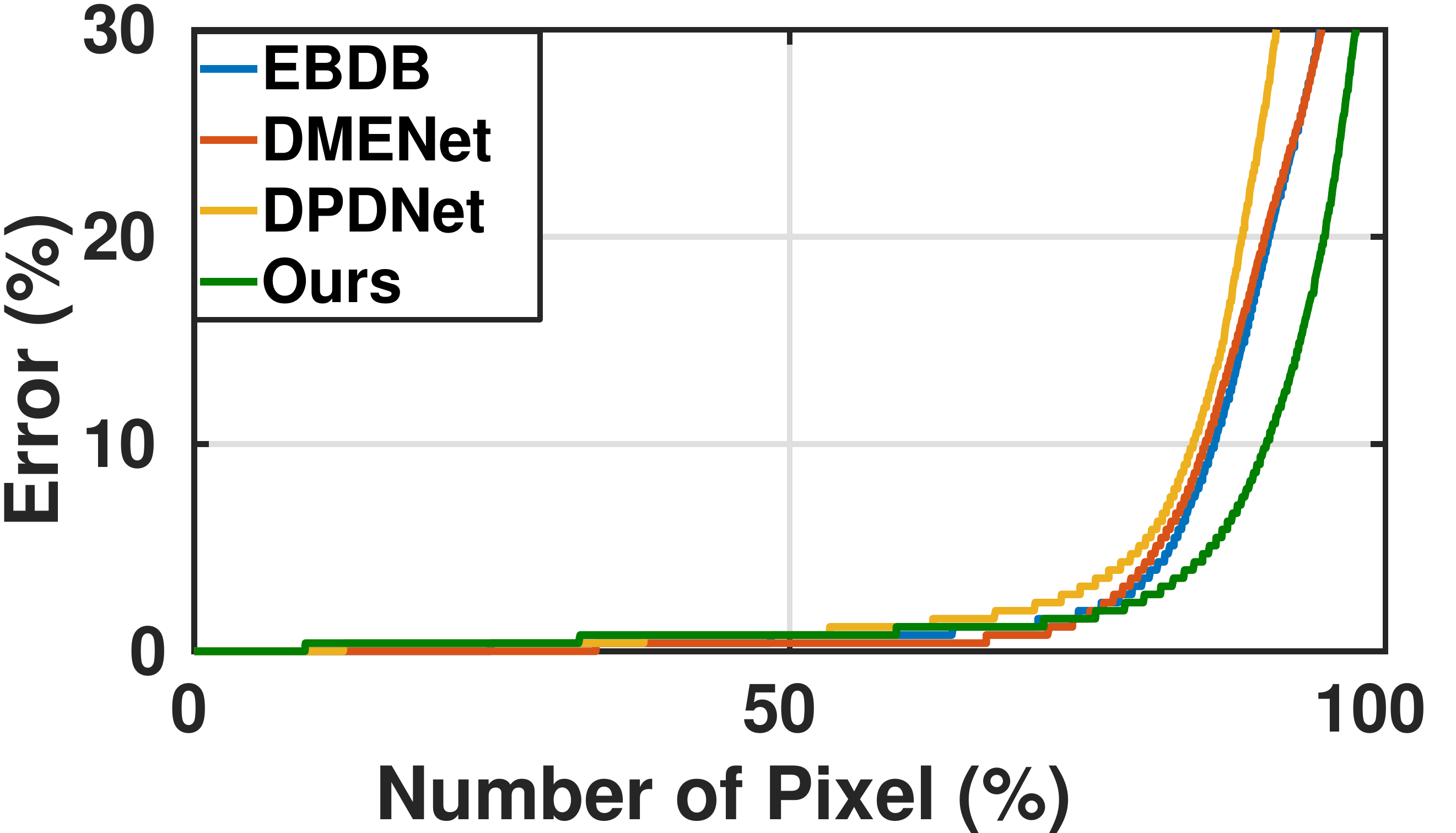} 
\hspace{-0.45 cm}
&\includegraphics[width=0.227\textwidth,height=0.1571
\textwidth]{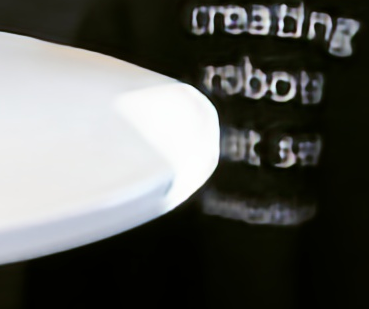} 
\hspace{-0.45 cm}
&\includegraphics[width=0.227\textwidth,height=0.1571
\textwidth]{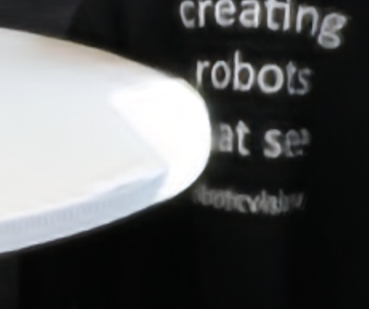}\\
\hspace{-0.25 cm}
\includegraphics[width=0.227\textwidth,height=0.1571
\textwidth]{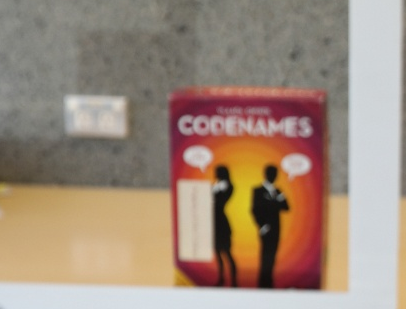}  
\hspace{-0.45 cm}
&\includegraphics[width=0.227\textwidth,height=0.1571
\textwidth]{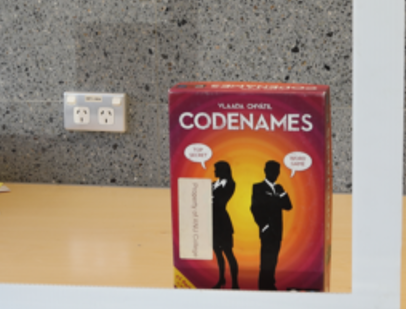} 
\hspace{-0.45 cm}
&\includegraphics[width=0.257\textwidth,height=0.1571
\textwidth]{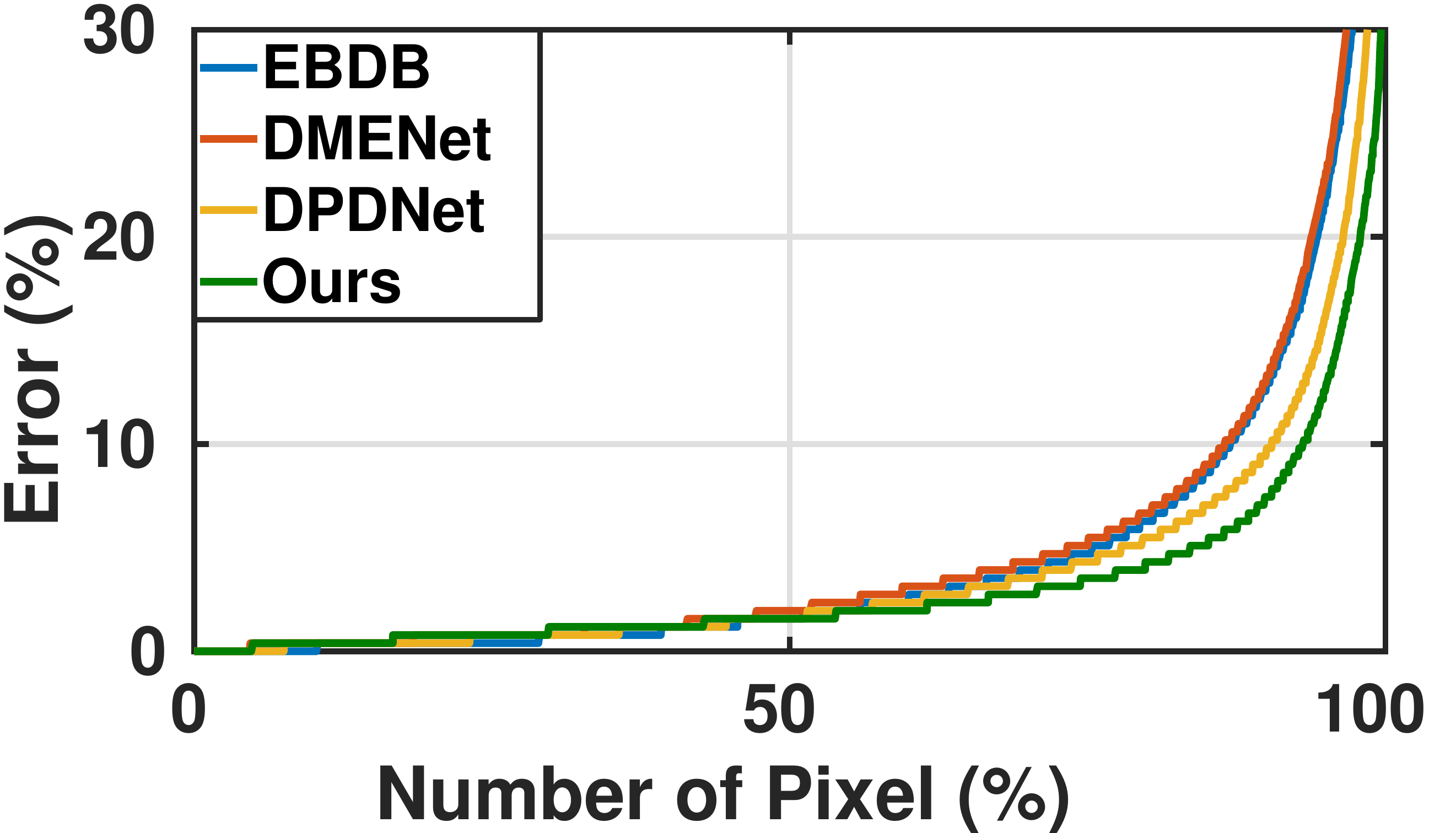} 
\hspace{-0.45 cm}
&\includegraphics[width=0.227\textwidth,height=0.1571
\textwidth]{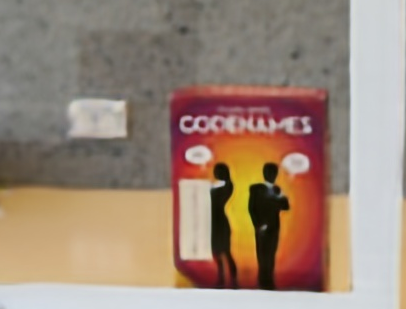} 
\hspace{-0.45 cm}
&\includegraphics[width=0.227\textwidth,height=0.1571
\textwidth]{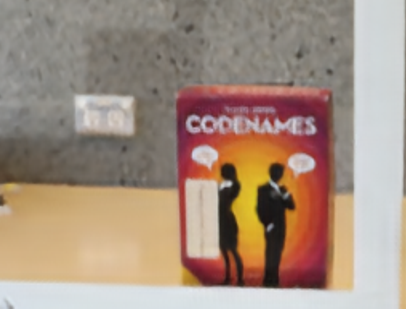}\\
\hspace{-0.25 cm}
\includegraphics[width=0.227\textwidth,height=0.1571
\textwidth]{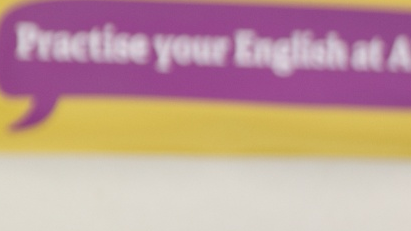}  
\hspace{-0.45 cm}
&\includegraphics[width=0.227\textwidth,height=0.1571
\textwidth]{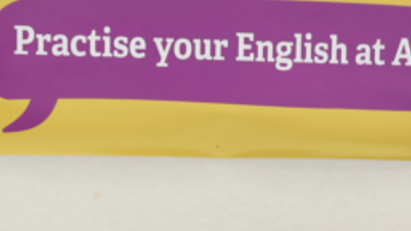} 
\hspace{-0.45 cm}
&\includegraphics[width=0.257\textwidth,height=0.1571
\textwidth]{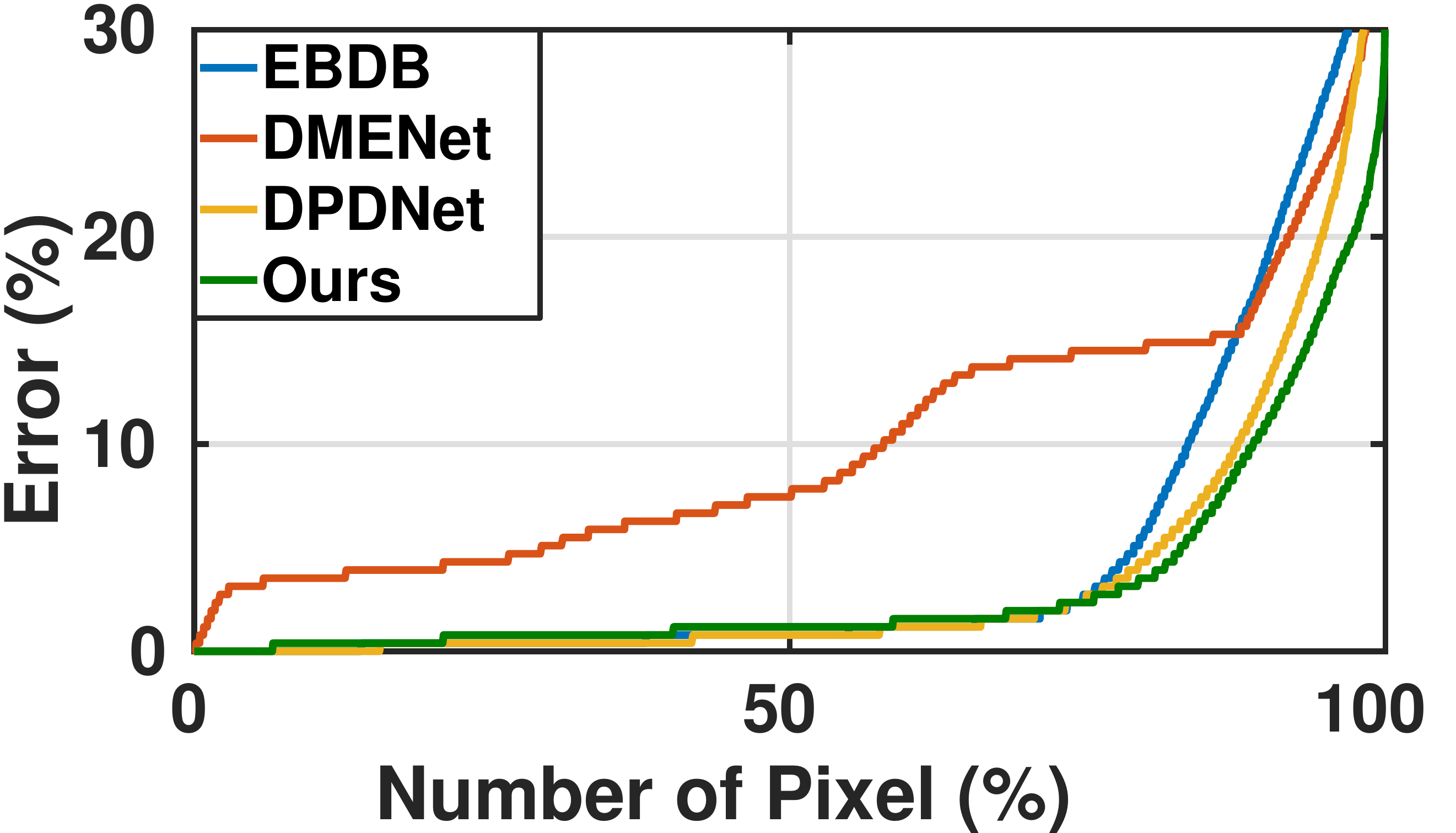} 
\hspace{-0.45 cm}
&\includegraphics[width=0.227\textwidth,height=0.1571
\textwidth]{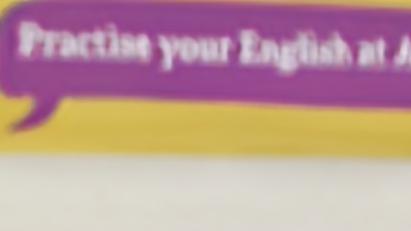} 
\hspace{-0.45 cm}
&\includegraphics[width=0.227\textwidth,height=0.1571
\textwidth]{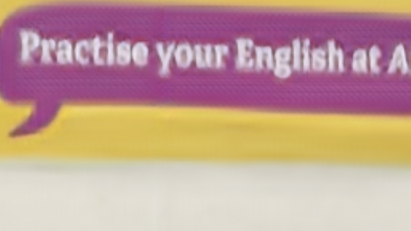}\\
\hspace{-0.25 cm}
(a) Input image
\hspace{-0.45 cm}
&(b) GT image
\hspace{-0.45 cm}
&(c) Intensity error
\hspace{-0.45 cm}
&(d) DPDNet~\cite{abuolaim2020defocus}
\hspace{-0.45 cm}
&(e) Ours: $\hat{\vI}$
\end{tabular}
}
\vspace{-2mm}
\end{center}
\caption{\label{fig:dpd-blur} \it
Examples of deblurring results on the real DPD-blur
dataset~\cite{abuolaim2020defocus} (top two rows) and our collected real DP
dataset (bottom three rows). (c) The distribution of the intensity error
(`RMSE\_rel'). It shows the percentile of the number of pixels below an error
ratio. The lower the curve, the better. For space limitations, we only display
deblurred images from the second-best baseline~\cite{punnappurath2020modeling}
for comparison. 
material. 
(Best viewed in colour on screen.)
}
\vspace{-3mm}
\end{figure*}

\begin{figure*}[t]
\begin{center}
\begin{tabular}{cccccc}
\hspace{-0.35 cm}
\includegraphics[width=0.169\textwidth]{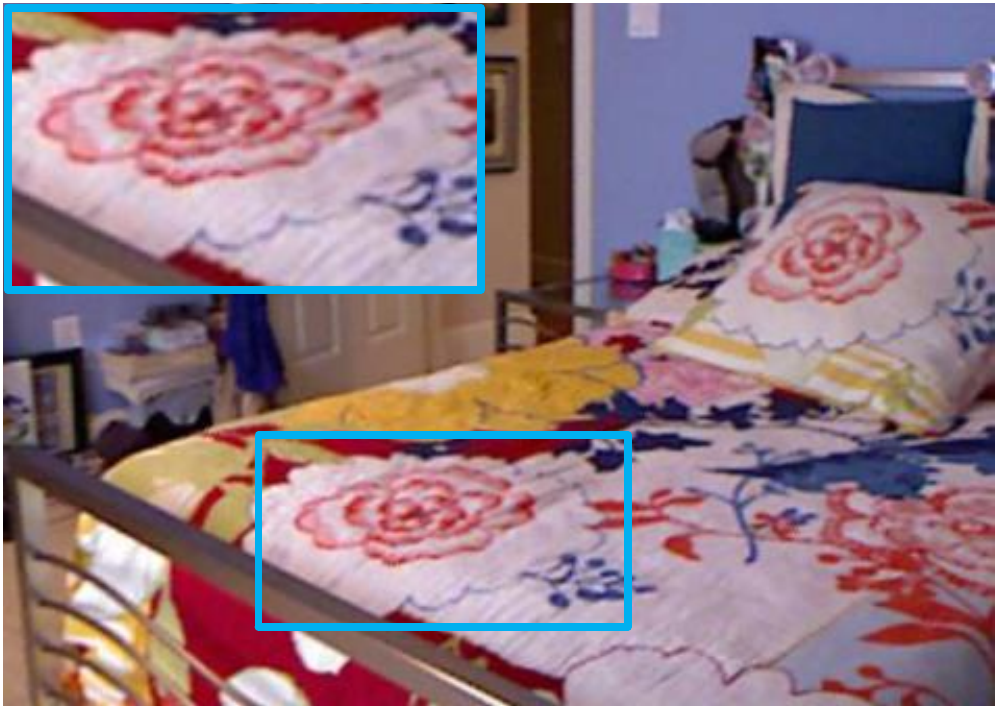}  
\hspace{-0.45 cm}
&\includegraphics[width=0.169\textwidth]{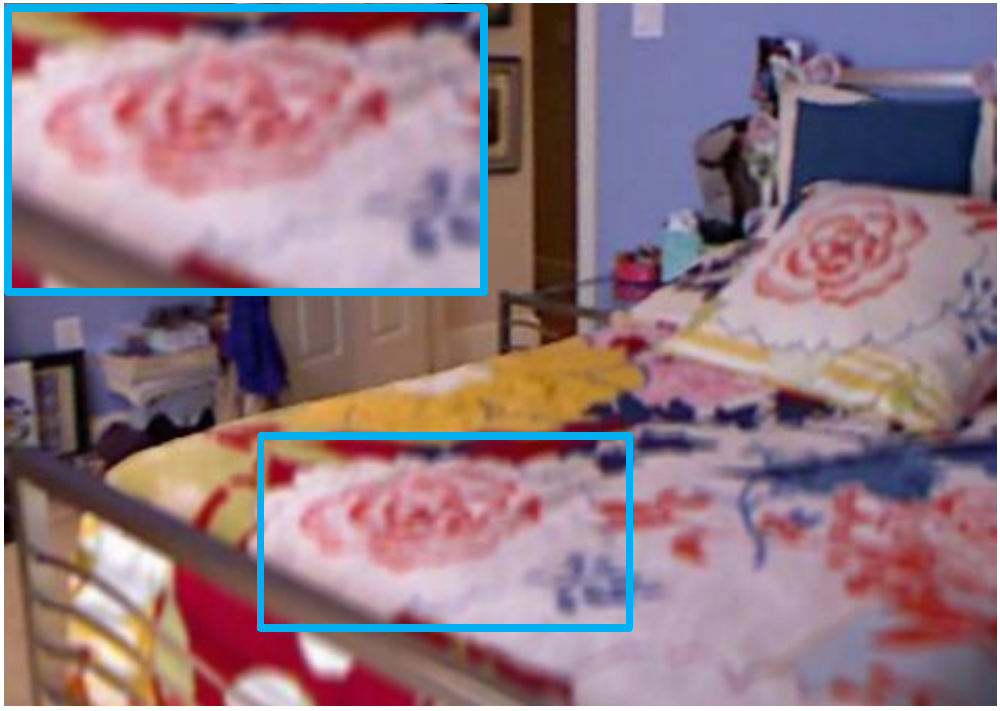}  
\hspace{-0.45 cm}
&\includegraphics[width=0.169\textwidth]{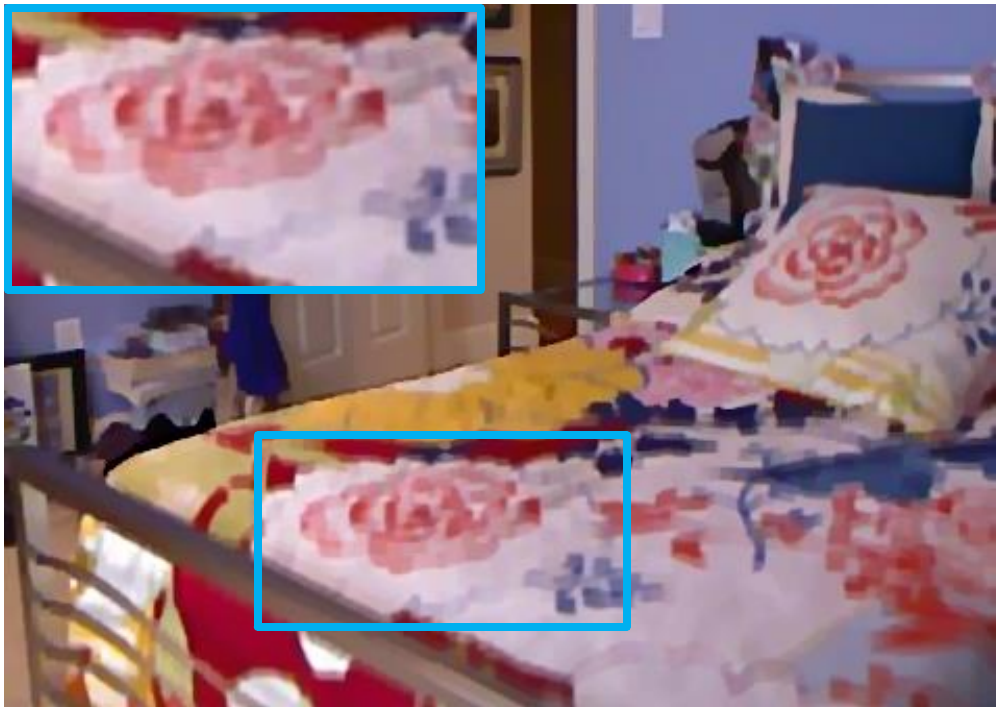}   
\hspace{-0.45 cm}
&\includegraphics[width=0.169\textwidth]{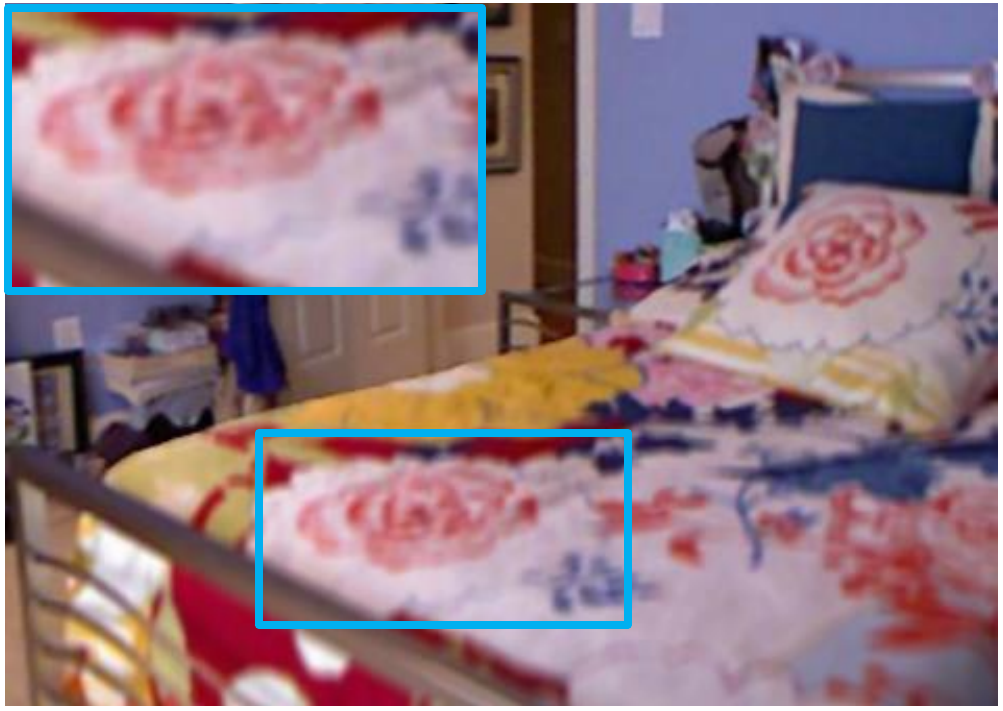}  
\hspace{-0.45 cm}
&\includegraphics[width=0.169\textwidth]{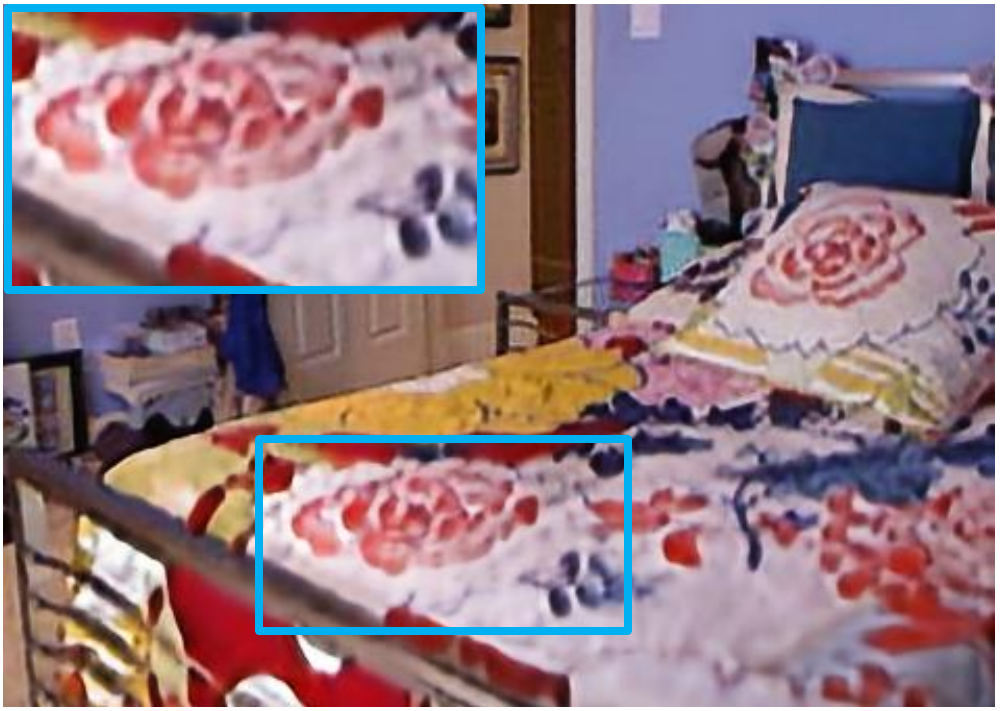} 
\hspace{-0.45 cm}
&\includegraphics[width=0.169\textwidth]{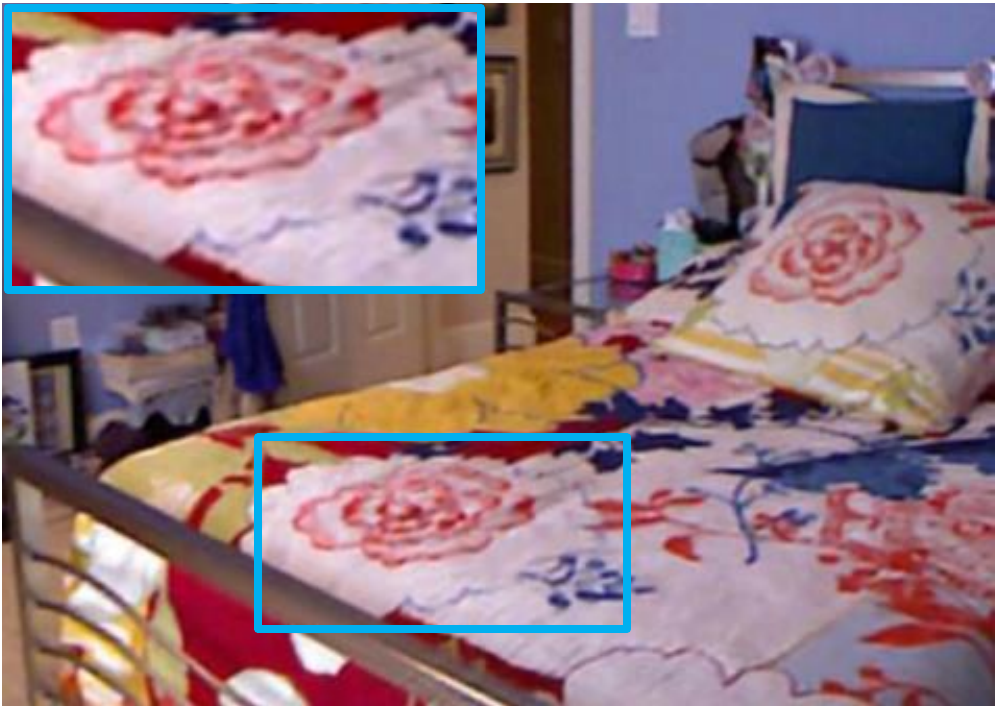}\\
\hspace{-0.35 cm}
(a) GT: Image $\vI$
\hspace{-0.45 cm}
&(b) Input image
\hspace{-0.45 cm}
&(c) EBDB \cite{karaali2017edge}
\hspace{-0.45 cm}
&(d) DMENet \cite{lee2019deep}
\hspace{-0.45 cm}
&(e) DPDNet \cite{abuolaim2020defocus}
\hspace{-0.45 cm}
&(f) Ours$_{wb}$: $\hat{\vI}$\\
\hspace{-0.35 cm}
\includegraphics[width=0.169\textwidth]{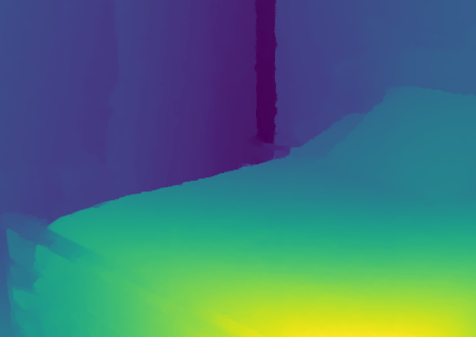} 
\hspace{-0.45 cm}
&\includegraphics[width=0.169\textwidth]{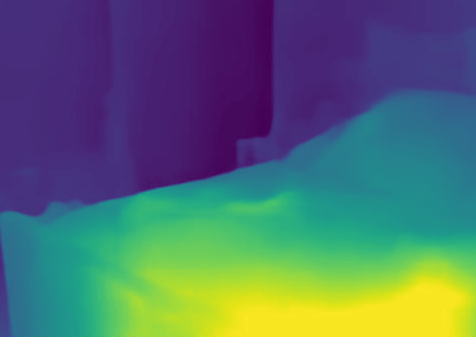}  
\hspace{-0.45 cm}
&\includegraphics[width=0.169\textwidth]{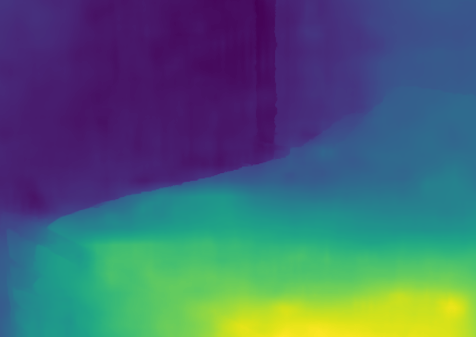}   
\hspace{-0.45 cm}
&\includegraphics[width=0.169\textwidth]{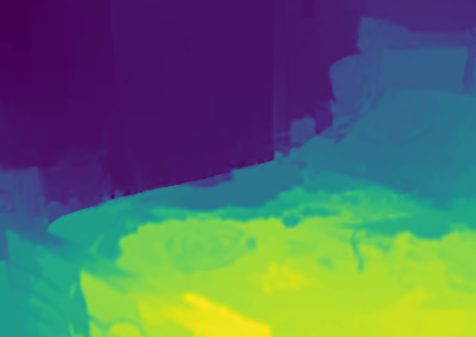} 
\hspace{-0.45 cm}
&\includegraphics[width=0.169\textwidth]{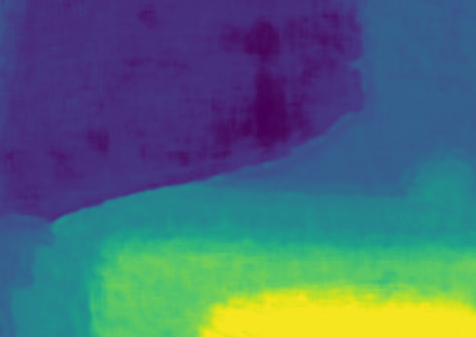}   
\hspace{-0.45 cm}
&\includegraphics[width=0.169\textwidth]{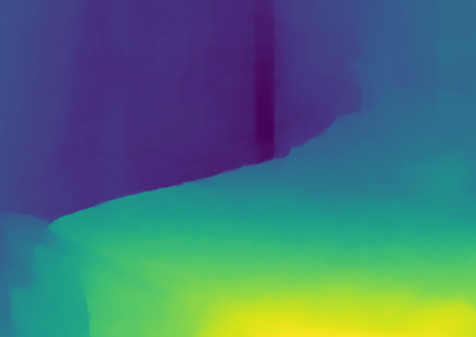}   \\
\hspace{-0.35 cm}
(g) GT: inverse depth $\vD$
\hspace{-0.45 cm}
&(h) BTS \cite{lee2019big} 
\hspace{-0.45 cm}
&(i) AnyNet \cite{wang2019anytime}
\hspace{-0.45 cm}
&(j) DPdisp \cite{punnappurath2020modeling}
\hspace{-0.45 cm}
&(k) Ours$_{b}$: $\hat{\vD}_c$
\hspace{-0.45 cm}
&(l) Ours$_{reb}$: $\hat{\vD}$\\
\end{tabular}
\begin{tabular}{cccc}
\hspace{-0.25 cm}
\includegraphics[width=0.43\textwidth,height=0.271
\textwidth]{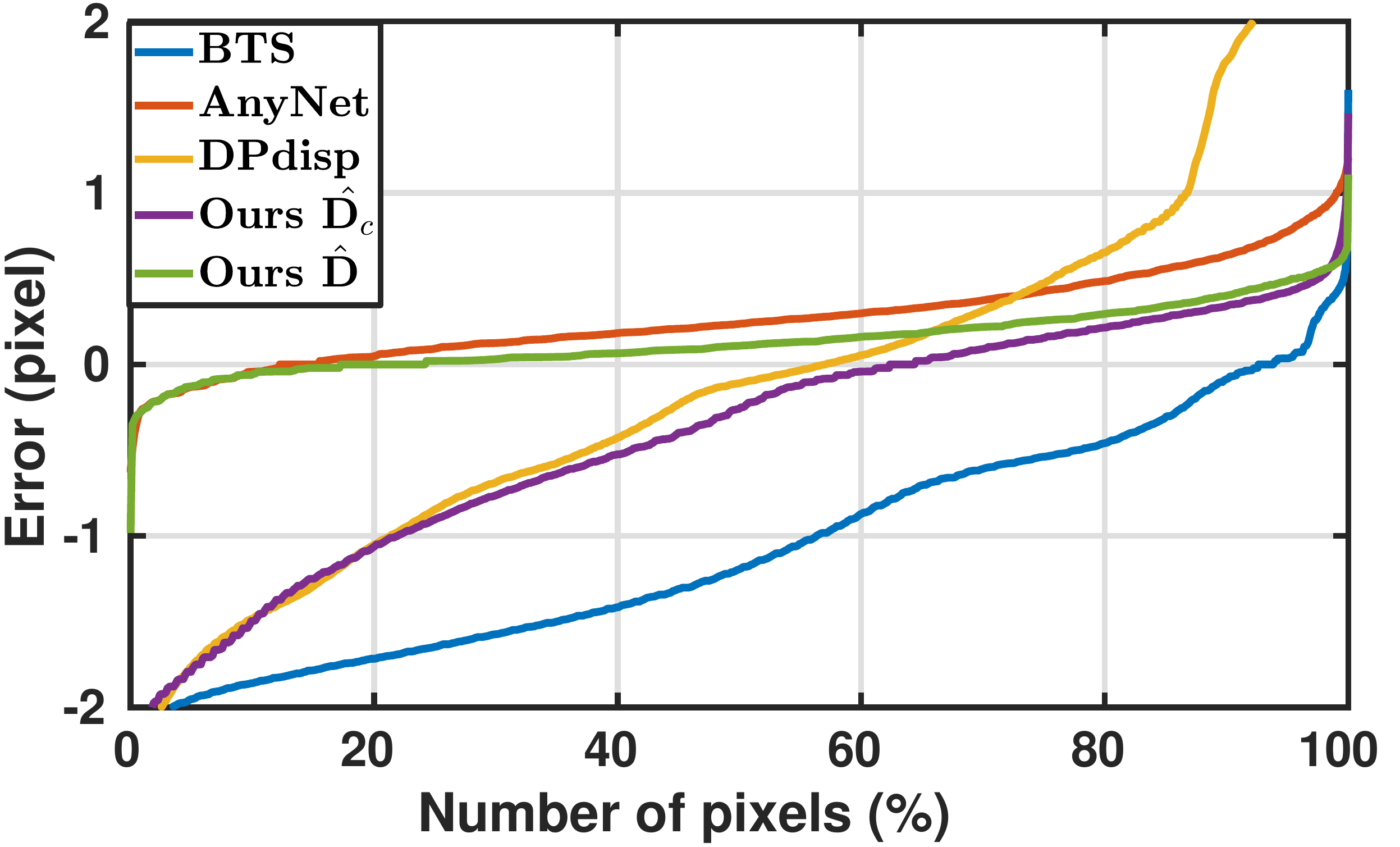}  
&\includegraphics[width=0.43\textwidth,height=0.271
\textwidth]{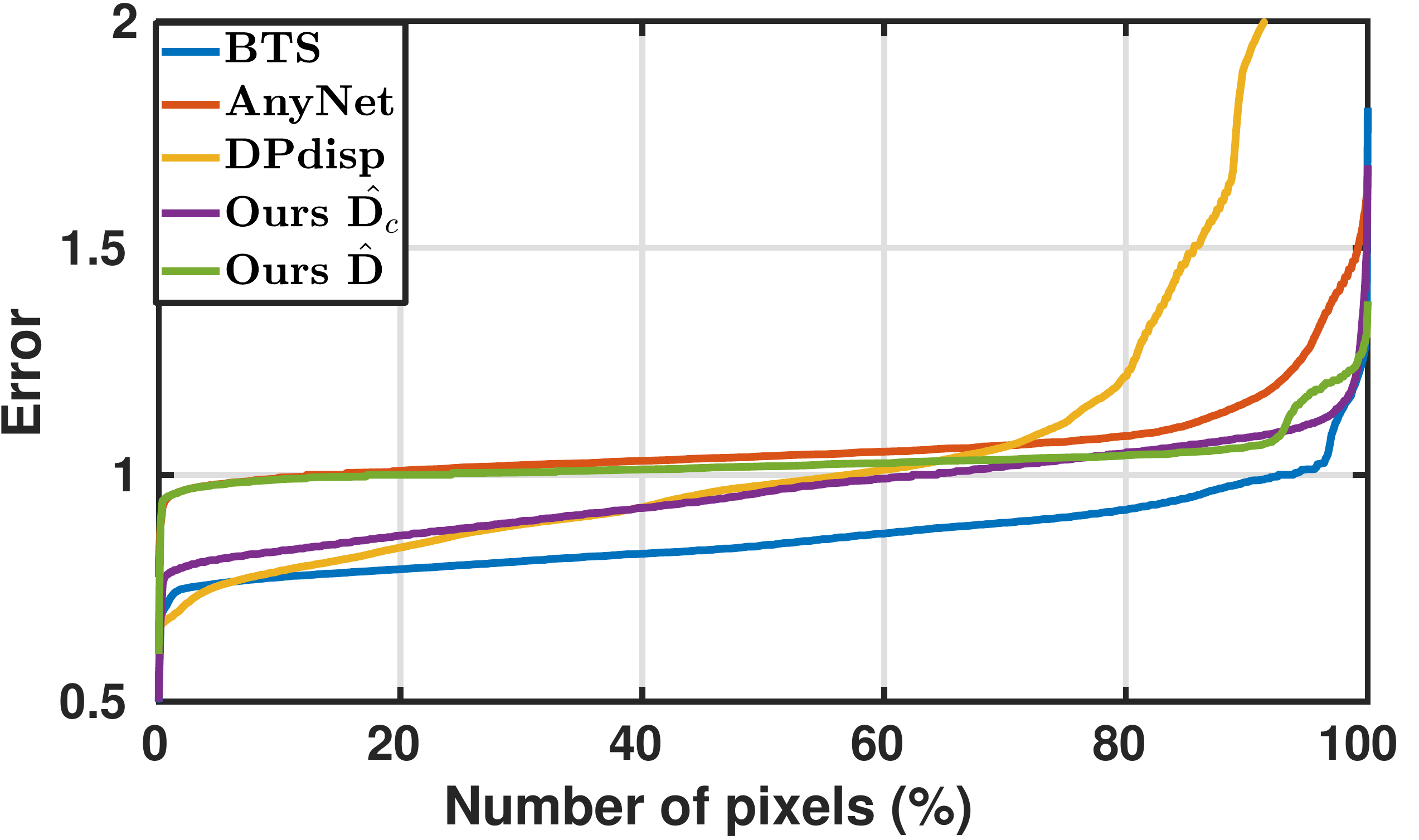} \\
\hspace{-0.25 cm}
(m) Inverse depth error, $\vD-\hat{\vD}$
&(n) Inverse depth error, $\vD/\hat{\vD}$\\
\end{tabular}
\end{center}
\vspace{-2mm}
\caption{\label{fig:oursyn} \it
An example of deblurring and inverse depth estimation results on our synthetic
dataset. 
(a) The ground truth image. 
(b) The left view of the simulated DP image pair. 
(c) Deblurred by EBDB~\cite{karaali2017edge}. 
(d) Deblurred by DMENet~\cite{lee2019deep}. 
(e) Deblurred by DPDNet~\cite{abuolaim2020defocus}. 
(f) Our deblurring result. 
(g) The ground truth inverse depth. 
(h) Depth by the monocular based method BTS~\cite{lee2019big} (converted to
inverse depth for display). 
(i) Inverse depth by the stereo based method AnyNet~\cite{wang2019anytime}. 
(j) Inverse depth by the DP based method DPdisp~\cite{punnappurath2020modeling}.  
(k) Output inverse depth of the  DepthNet (see Fig.~\ref{fig:pipeline}). 
(l) Our inverse depth result. 
(m) Inverse depth error: $\vD-\hat{\vD}$, the closer to zero, the better.
(n) Inverse depth error: $\vD/\hat{\vD}$, the closer to $1$, the better. The error
distributions help to statistically analyze the accuracy of each method. Our
method outperforms other baselines significantly. (Best viewed in colour on
screen.)
}
\vspace{-2mm}
\end{figure*}
\subsection{Results and discussions}

We compare our results with baselines on depth estimation and image deblurring
on 4 (including real and synthetic) datasets.

\vspace*{+0.2 mm}
\noindent{\bf{Deblurring results.}}
We show the quantitative and qualitative comparisons for deblurring in
Table~\ref{tab:Deblur-table},  Fig.~\ref{fig:dpd-blur} and
Fig.~\ref{fig:oursyn}, respectively.
In Table~\ref{tab:Deblur-table}, we achieve competitive results compared with
the state-of-the-art methods~
\cite{karaali2017edge,lee2019deep,abuolaim2020defocus} on both DPD-blur, Our-syn, and Our-real dataset.~Here,
`Ours$_{\rm wb}$' and `Ours$_{\rm reb}$' denote results without and with the
reblur loss, respectively. The relative improvement is around $10\%$,
demonstrating that using our DP-model based reblur loss can significantly
improve the deblurring performance of our DDDNet.  
Fig.~\ref{fig:dpd-blur} and Fig.~\ref{fig:oursyn} show the qualitative and
quantitative comparisons on the three datasets. 

\vspace*{+0.2 mm}
\noindent{\bf Depth results.} 
We first provide depth estimation results on the DPD-disp dataset. The
comparisons are shown in Table \ref{tab:DPD-disp} and Fig.~\ref{fig:dpd-disp}.
As no training data is available, we directly test our model (trained on Our-syn
dataset) on their testing set without fine-tuning and achieve the second best
performance. Furthermore, note that our DP-model based reblur loss can be used
as a `self-supervision' signal to adapt our DDDNet to this test dataset, we can
further fine-tune our model trained on Our-syn dataset using our reblur loss,
without using GT depth map. 
Our model with reblur loss `Ours (fine-tune)' achieves competitive results
compared with the state-of-the-art methods
\cite{lee2019big,monodepth2,wadhwa2018synthetic,punnappurath2020modeling}.  

We then provide depth comparisons on Our-syn dataset. The results are shown in
Table~\ref{tab:syn} and Fig.~\ref{fig:oursyn}. In Table~\ref{tab:syn}, we
evaluate the outputs of our DDDNet step by step, it shows that, with our joint
optimization of depth estimation and deblurring, we improve the quality of both
the depth map and the  deblurred image. Also, with our DP-model based reblur
loss, we further improve our performance. For example, for the `abs\_rel'
metric, the relative error for the coarse depth map as the output of the
DepthNet (see Fig.~\ref{fig:pipeline}), the DeblurNet without reblur loss
and the DeblurNet with reblur loss is $0.149$, $0.091$ and $0.083$,
respectively.


\noindent{\bf Effectiveness of our simulated DP images.}
To further show the usefulness of our simulated DP images, our model trained on
Our-syn dataset is directly used to test on the DPD-blur dataset.~The result is
$20.28$dB/$0.650$/$9.68\%$ in PSNR/SSIM/RMSE\_rel without any fine-tuning.~After
fine-tuning, our model trained on Our-syn get remarkable performance, with
PSNR/SSIM/RMSE\_rel at $26.92$dB/$0.864$/$4.51\%$, outperforming the performance
of our model ($26.76$dB/$0.891$/$4.59\%$) directly trained on the DPD-blur
dataset. To further show the transfer ability of our model trained using
simulated DP images, we only use a half amount of images from the DPD-blur
dataset to finetune our model trained on Our-syn dataset, obtaining a good
performance, with PSNR/SSIM/RMSE\_rel at $26.52$dB/$0.822$/$4.72\%$ in $20$
epoches.   

\vspace{-0.5mm}
\section{Conclusions}
\vspace{-0.5mm}

In this paper, we derive a mathematical DP model formulating the imaging process
of the DP camera. We propose an end-to-end \textbf{DDDNet} (DP-based Depth
and Deblur Network), to jointly estimate a inverse depth map and restore a sharp
image from a blurred DP image pair. Also, we present a reblur loss building upon
our DP model and integrate it into our DDDNet. For future work, we plan to
extend our DDDNet to a self-supervised depth and sharp image estimation
network.

{\small
\bibliographystyle{ieee_fullname}
\bibliography{arxiv-2021}
}

\end{document}